\documentclass[11pt]{article}

\usepackage[final]{acl}

\usepackage{times}
\usepackage{latexsym}

\usepackage[T1]{fontenc}

\usepackage[utf8]{inputenc}

\usepackage{microtype}

\usepackage{inconsolata}

\usepackage{graphicx}

\usepackage{multirow}
\usepackage{subfigure}
\usepackage{amsmath}
\usepackage{amssymb}
\usepackage{pifont}
\usepackage{graphicx}
\usepackage{color}
\usepackage{longtable}
\usepackage{tcolorbox}

\title{Pattern Over-Generalization of Knowledge Graph Embedding}

\author{Junsik Kim \and Kangil Kim\thanks{\; Corresponding author.}
\\
AI Graduate School \\
Gwangju Institute of Science and Technology \\
  \texttt{junsikkim@gm.gist.ac.kr},
  \texttt{kangil.kim.01@gmail.com} \\}

\begin{document}
\maketitle
\begin{abstract}
Knowledge graph embedding (KGE) demonstrates its effectiveness for predicting missing links in knowledge graphs (KGs) by projecting entities and relations into a low-dimensional vector space. It is crucial for KGE models to effectively capture inference patterns (patterns) inherent in KGs, such as symmetry/antisymmetry, inversion and composition. Although recent KGE models exhibit strong capabilities in modeling such diverse patterns, they suffer from inherent limitations stemming from \textit{pattern over-generalization}, where embeddings learned from only a single pattern instance inevitably generalize that pattern to all related instances, i.e., generalize the pattern universally. To address this issue, we propose PogRE (Pattern Over-Generalization Robust Embedding), a simple but effective method that utilizes dense linear transformations and compound operations for relation representation. Our theoretical analysis demonstrates that a dense linear transformation allows a pattern to become progressively universal as more triples are observed in the pattern. Furthermore, after observing  $d+1$ linearly independent entities ($d+1$ denotes the dimension of entity), the linear transformation guarantees universal generalization of the pattern across all related instances. Experimental results on three standard benchmark datasets show that PogRE outperforms existing state-of-the-art KGE models in link prediction. Moreover, our empirical results indicate that PogRE effectively addresses the negative impact of over-generalization.
\end{abstract}

\section{Introduction}
Knowledge graphs (KGs) store vast amounts of human knowledge in the form of triples $(h, r, t)$, where $h$ and $t$ represent the head and tail entities and $r$ denotes the relationship between entities. KGs have demonstrated their effectiveness in various downstream tasks~\cite{sui2025fidelis, ma-etal-2025-large-language-models-meet}. However, real-world KGs such as Freebase~\cite{bollacker2008freebase} and WordNet~\cite{miller1995wordnet}, even on a large scale, still suffer from incompleteness~\cite{bordes2013translating}. To address this issue, Knowledge graph embedding (KGE), which represents entities and relations in a low-dimensional vector space, has been widely studied as an effective method for predicting missing links.

A fundamental challenge in KGE lies in effectively capturing the inference patterns (patterns) inherent in KGs, such as symmetry/antisymmetry, inversion, and composition. To address this, existing works focus on designing specific score functions to capture these patterns. For instance, TransE~\cite{bordes2013translating} represents relations as translations to model inversion and composition, while RotatE~\cite{sun2019rotate} employs rotations to capture symmetry/antisymmetry, inversion and composition. PairRE~\cite{chao2021pairre} and CompoundE~\cite{ge2023compounding} leverage scaling and compound operators to effectively model more patterns, including subrelation as well as complex relations.

Despite their strong ability to capture various patterns, existing KGE models tend to over-generalize the patterns they observe. In particular, once a model observes a pattern, it generalizes the pattern universally across the entire graph, even when the pattern is supported by only a small number of observed triples. Consequently, patterns that are valid locally in the graph are treated as universally valid. We refer to this phenomenon as over-generalization, which leads to erroneous predictions.

To address this issue, we propose a simple but effective method, PogRE, that prevents locally valid patterns from being generalized universally. PogRE uses dense linear transformations and compound operations for relation representation, where the linear transformation is decomposed into a relation-specific orthogonal matrix and a shared upper-triangular matrix. This framework theoretically guarantees that patterns supported by only a small number of observed triples are generalized locally, while ensuring that any pattern supported by sufficient observed triples is generalized universally across the entire graph, when patterns are represented as connected relational paths.

Our contributions are as follows:
\begin{itemize}
\item We introduce pattern over-generalization, the phenomenon in which patterns supported by only a small number of observed triples are generalized universally across the entire graph in existing KGE models.
\item We propose a novel KGE method, Pattern Over-Generalization Robust Embedding (PogRE), and theoretically guarantee that it effectively addresses over-generalization. In particular, PogRE allows any pattern represented as a connected relational path to become progressively universal as more triples are observed in the pattern. Moreover, sufficient triples are observed, PogRE guarantees universal generalization of the pattern across all related instances. 
\item Experimental results on three benchmark datasets demonstrate that PogRE consistently outperforms baseline KGE models in link prediction and effectively addresses the negative impact of over-generalization.
\end{itemize}

\section{Background}
\paragraph{Knowledge Graph Embedding}
Given sets of entities and relations $E$ and $R$, a KG can be defined as a collection of factual triples $G= \{(h, r, t)| h, t \in E, r \in R\}$, where $h$ and $t$ are the head and tail entities, and $r$ is the relation. KGE maps $E$ and $R$ to low-dimensional vector space and defines a score function to measure triple plausibility.

Distance-based models (DBMs) are trained to minimize the distance of the factual triple $(h, r, t)$, while maximizing the distance of corrupted negative triples $(h', r, t)$ or $(h, r, t')$, which are generated by randomly replacing the head $h$ or tail $t$ with other entities in $E$. PairRE~\cite{chao2021pairre}, a representative model of DBMs, defines the score function as follows: 
\begin{equation}
\begin{aligned}
\textstyle f_r(h, t) = \|h\circ r^H - t \circ r^T\|, 
\label{eq:Equation_1}
\end{aligned}
\end{equation}

where $h, t, r^H, r^T \in \mathbb{R}^d$, $\circ$ denotes a Hadamard product and $\| \cdot \|$ is a vector norm.

Tensor decomposition models (TDMs) are trained to maximize the score (or semantic similarity) of the factual triple calculated via the multi-linear product of the head entity $h$, the relation $r$ and the tail entity $t$, while minimizing the score of negative triples. DistMult~\cite{yang2015embedding}, a representative model of TDMs, defines the score function as follows: 
\begin{equation}
\begin{aligned}
\textstyle f_r(h, t) = \langle h, r, t \rangle = \sum_{i=1}^{d} h_i r_i t_i,
\end{aligned}
\end{equation}
where $h, r, t \in \mathbb{R}^d$, and $\langle \cdot, \cdot, \cdot \rangle$ denotes the sum of element-wise products.

\paragraph{Inference Pattern}
Inference patterns (patterns) are widely used to analyze the generalization capabilities of KGEs. 
A pattern, notated as $\psi \Rightarrow \phi$, has the body $\psi$ and the head $\phi$, which are sets of triples composed of observed entities and relations in the data. 
For example, a composition pattern for relations $r_1, r_2, r_3\in R$ is defined as  $r_1(X,Y) \wedge r_2(Y, Z) \Rightarrow r_3(X,Z)$. 
A pattern implies that \textit{if the body is in the graph, the head is also in the graph}~\cite{pavlovicexpressive}.

\section{Problem} 
\label{sec:problem}

\subsection{Problem Formulation}
\paragraph{Pattern Instance}
We further define a pattern instance as an instantiation of pattern $\psi \Rightarrow \phi$. For the composition pattern of relations $r_1, r_2, r_3 \in R$, a pattern instance $\psi_1 \Rightarrow \phi_1$ is expressed as $r_1(e_x, e_y) \wedge r_2(e_y, e_z) \Rightarrow r_3(e_x, e_z)$, where $r_1(e_x, e_y) \wedge r_2(e_y, e_z)$ corresponds to a body instance $\psi_1$, $r_3(e_x, e_z)$ corresponds to the head instance $\phi_1$, and $e_x, e_y, e_z \in E$.

\paragraph{Pattern Over-Generalization}
Although well-known KGE models such as TransE~\cite{bordes2013translating}, RotatE~\cite{sun2019rotate}, PairRE~\cite{chao2021pairre}, and CompoundE~\cite{ge2023compounding} demonstrate strong generalization capabilities by modeling various patterns, they suffer from inherent limitations stemming from \textit{pattern over-generalization}. Pattern over-generalization is the phenomenon where a model, after observing a single instance of a pattern ($\psi_1 \Rightarrow \phi_1$) in the graph $G$, generalizes the pattern to every body instance that appears in the graph; for example, if the body $\psi_2$ appears in graph $G$, the model infers that the corresponding head $\phi_2$ must also exist (i.e., $\psi_2 \Rightarrow \phi_2$). This issue arises because existing models are trained to generalize a pattern universally. We formally define the phenomenon of \textit{pattern over-generalization} as well as \textit{local pattern}, and \textit{universal pattern} as follows:

\begin{tcolorbox}[
    colback=blue!5!white,
    colframe=blue!75!black,
    title=Pattern Over-Generalization,
    boxsep=4pt,
    left=2pt,
    right=2pt,
    top=2pt,
    bottom=2pt
]

The model generalizes a \textit{local pattern} to all unseen triples without sufficient evidence, i.e., the model treats a \textit{local pattern} as a \textit{universal pattern}.

\vspace{0.3em} 
\noindent \textbullet~Local Pattern: $\psi_i \in G_{o} \Rightarrow \phi_i \in G_{p}$ (s.t. $G_p \subset G_u$ and $G_p \neq G_u$) \\
\noindent \textbullet~Universal Pattern: $\psi_i \in G_{o} \Rightarrow \phi_i \in G_{p}$ (s.t. $G_p = G_u$)

\noindent for a given set of instantiated relations $\{r_i\}_{i=1}^n$.

\end{tcolorbox}

where
\begin{equation}
\begin{aligned}
G_{o}:&\ \{f|f \text{ is a set of triples observed in the given} \\ & \qquad  \text{KG}\} \\
G_{p}:&\ \{f|f \text{ is } \psi_i \in G_{o} \text{ or } \phi_i \text{ that corresponds to } \psi_i, \\ & \qquad \text{ s.t. } \phi_i \text{ is semantically correct.}\} \\
G_{u}:&\ \{f|f \text{ is } \psi_i \in G_{o} \text{ or } \phi_i \text{ that corresponds to } \psi_i)\}
\end{aligned}
\end{equation}

While this phenomenon can serve as a crucial inductive bias in KGE, universally generalizing a local pattern without sufficient evidence can lead to erroneous predictions by injecting incorrect information into the embeddings.

\subsection{Cause and Evidence}
\paragraph{Why Does The Problem Appear?}

The cause is that the pattern condition depends only on relation embeddings in KGE methods. This entity-independent pattern condition allows the model to generalize local patterns to unseen triples.

For example, in PairRE that is fully trained to satisfy Equation~\ref{eq:Equation_1} for all triples in $G$, if body instance $(e_{x_1}, r_1, e_{y_1}), (e_{y_1}, r_2, e_{z_1}) \in G$ and head instance  $(e_{x_1}, r_3, e_{z_1}) \in G$, we have 
\begin{equation}
\begin{aligned}
\textstyle e_{x_1} \circ r_1^{H} = e_{y_1} \circ r_1^{T} 
\;\wedge\;
e_{y_1} \circ r_2^{H} = e_{z_1} \circ r_2^{T} \\
\textstyle \;\wedge\;
e_{x_1} \circ r_3^{H} = e_{z_1} \circ r_3^{T} \\
\textstyle \Rightarrow\quad
r_1^{T} \circ r_2^{T} \circ r_3^{H}
=
r_1^{H} \circ r_2^{H} \circ r_3^{T}
\end{aligned}
\label{eq:pattern equation}
\end{equation}

\begin{figure}[t]
    \centering
    \includegraphics[width=1\columnwidth]{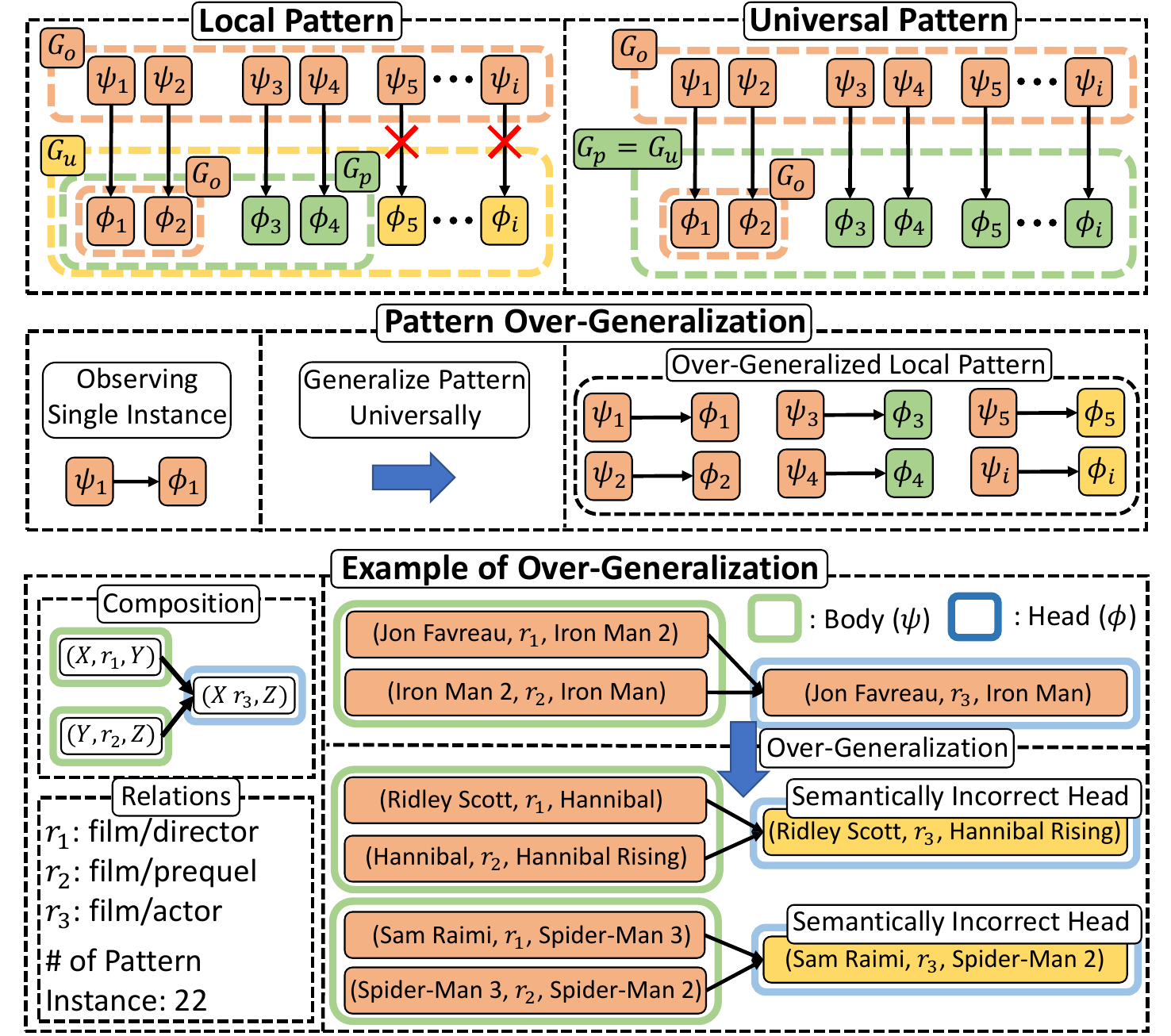}
    \caption{Illustration of local and universal patterns, and the process and examples of over-generalization. Existing models suffer from over-generalization by treating local patterns as universal patterns. The example shows that a local pattern is generalized universally, which leads to erroneous predictions.}
    \label{fig:method_overview}
\end{figure}

Under this pattern condition, if a new body instance $(e_{x_2}, r_1, e_{y_2}), (e_{y_2}, r_2, e_{z_2})$ is observed as
\begin{equation}
\begin{aligned}
\textstyle e_{x_2} \circ r_1^{H} = e_{y_2} \circ r_1^{T} 
\;\wedge\;
e_{y_2} \circ r_2^{H} = e_{z_2} \circ r_2^{T}
\end{aligned}
\end{equation}

then the model guarantees that
\begin{equation}
\begin{aligned}
\textstyle e_{x_2} \circ r_3^{H} = e_{z_2} \circ r_3^{T}, 
\end{aligned}
\end{equation}
thereby leading the model to treat the corresponding head as valid for every new body instance of the same pattern.
This phenomenon is further illustrated in Figure~\ref{fig:method_overview}, where the model generalizes the pattern universally to every body instance. However, not all patterns in KGs are universally valid, especially those with a few pattern instances.\footnote{In Appendix~\ref{appendix:empirical_local_global}, we present examples of semantically local patterns that have low frequency, along with failure cases where pattern frequency does not match the pattern semantics.}

\begin{figure}[t]
    \centering
    
    \subfigure[Histograms of local patterns that are supported by scarce pattern instances. The relation sets for the left and right figures are \textit{(film/written\_by, actor/film, film/prequel)} and \textit{(film/director, film/prequel, actor/film)}, respectively]{
        \includegraphics[width=0.5\columnwidth]{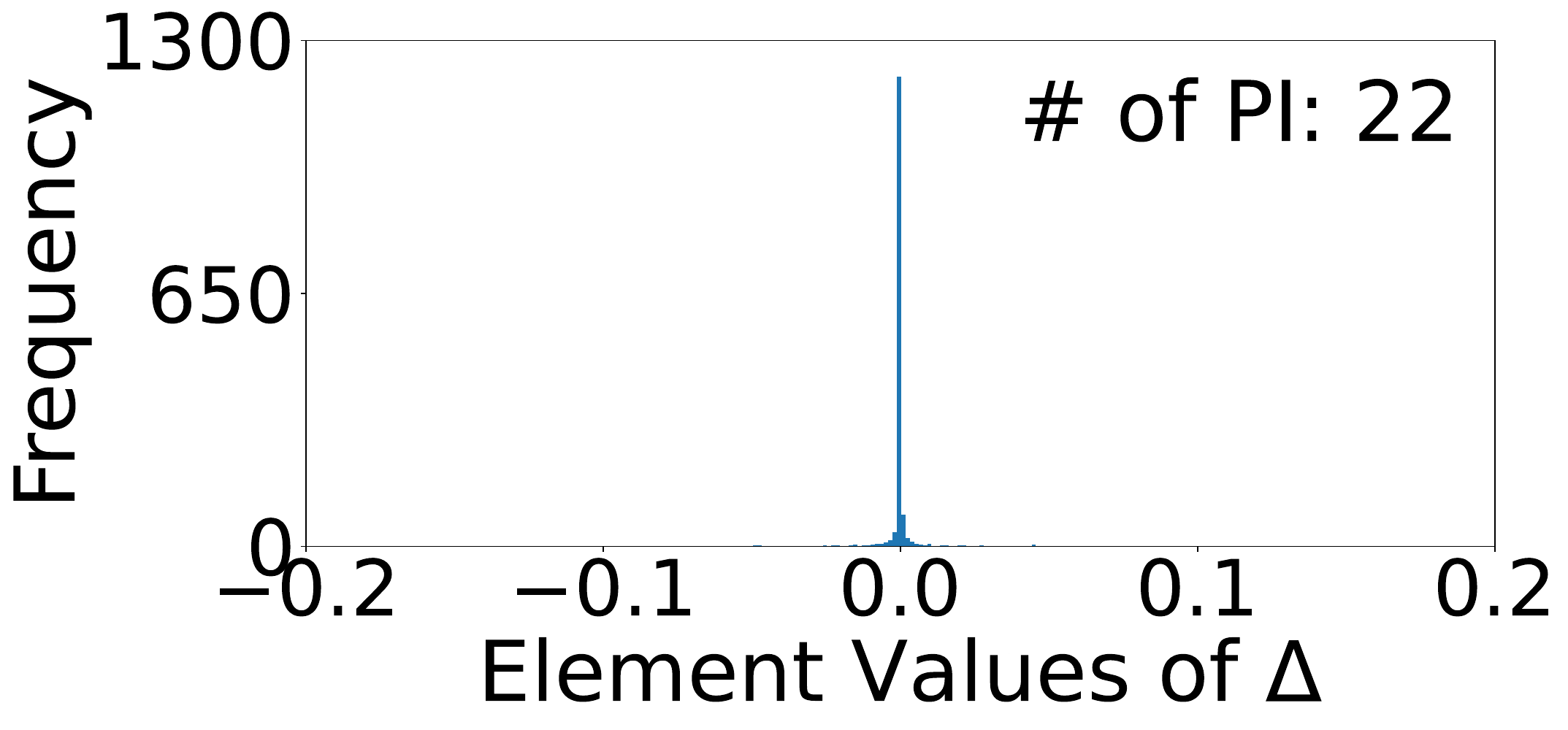}
        \hfill
        \includegraphics[width=0.5\columnwidth]{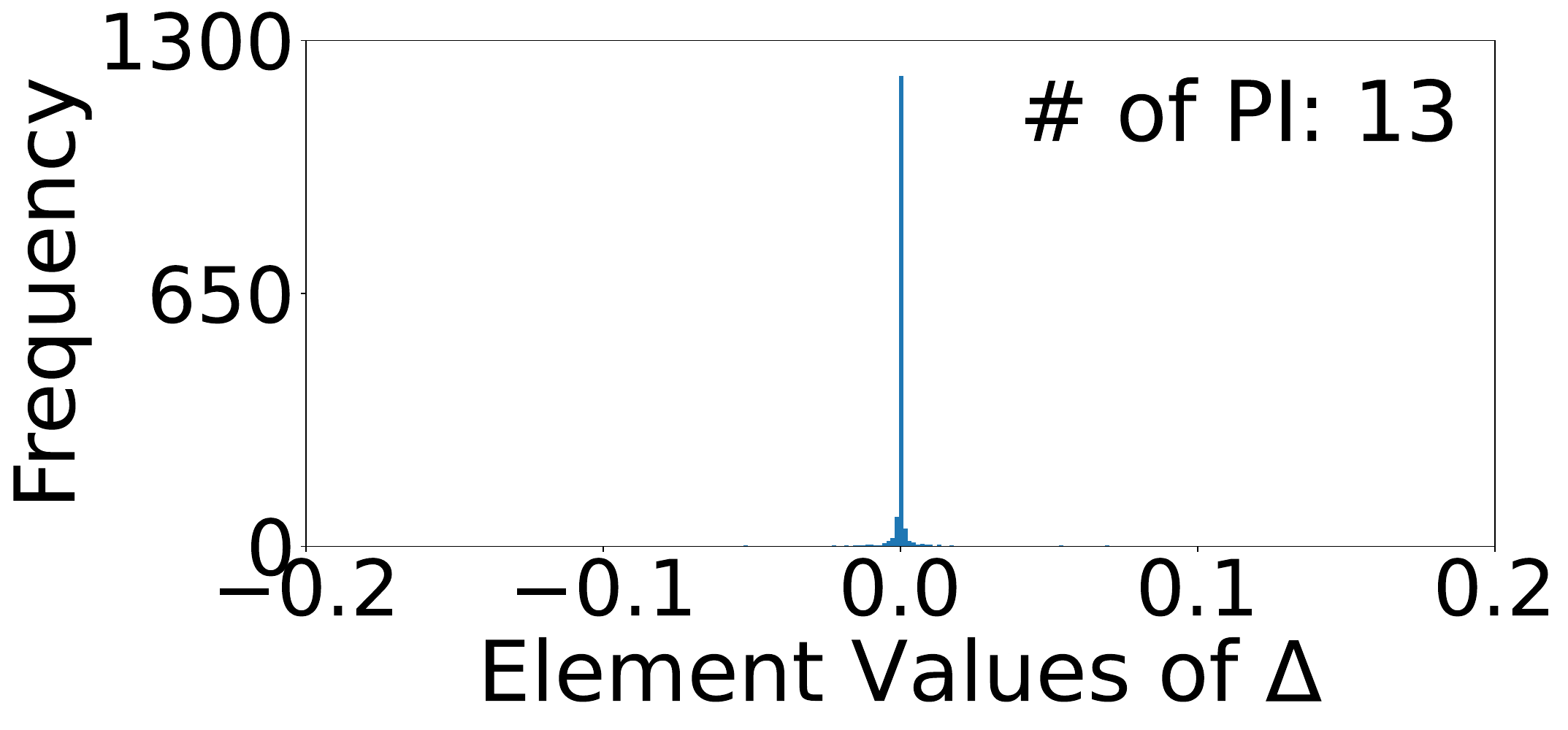}
        \label{fig:problem a}
    }
    
    \vspace{1em} 
    
    \subfigure[Histograms of universal patterns that are supported by many pattern instances. The relation sets for the left and right figures are \textit{(actor/film, film/country, people/nationality)} and \textit{(people/place\_of\_birth, location/country, people/nationality)}, respectively]{
        \includegraphics[width=0.5\columnwidth]{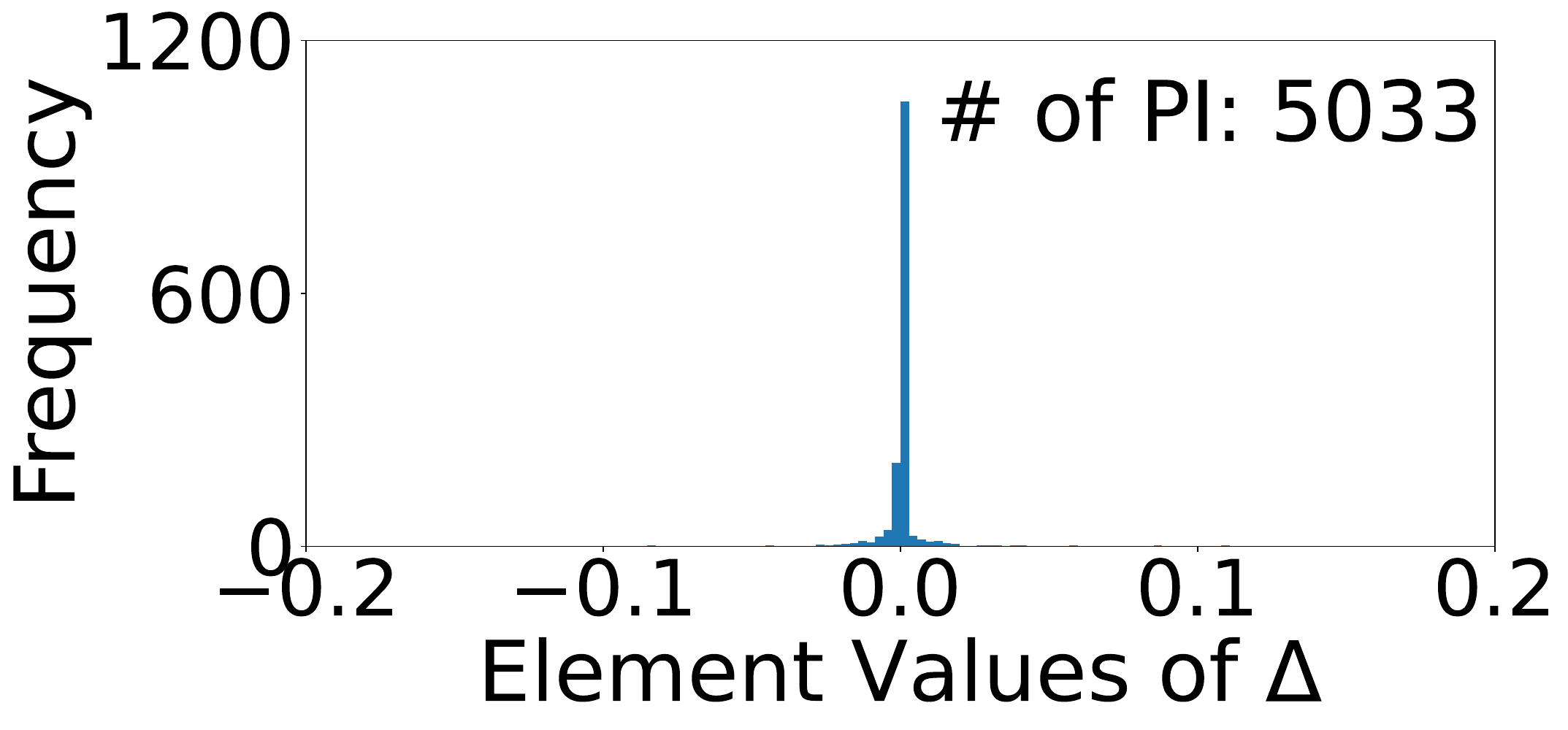}
        \hfill
        \includegraphics[width=0.5\columnwidth]{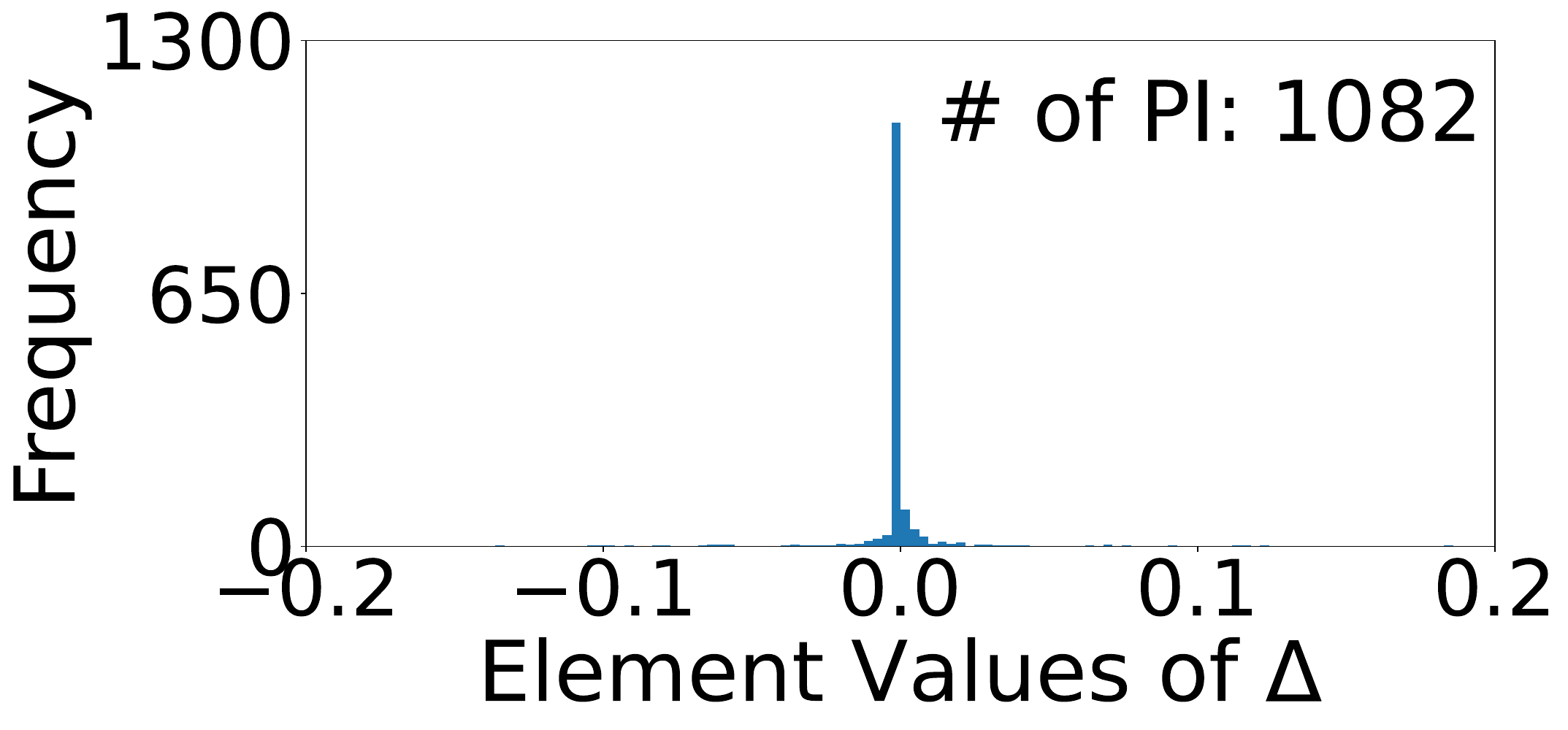}
        \label{fig:problem b}
    }
    
    \caption{Histograms of embedding difference $\Delta =r_1^{T} \circ r_2^{T} \circ r_3^{H} - r_1^{H} \circ r_2^{H} \circ r_3^{T}$ for different relation set $(r_1, r_2, r_3)$. \# of PI denotes the number of pattern instances. $(r_1, r_2, r_3)$ are retrieved from FB15k-237.}
    \label{fig:problem}
\end{figure}

\begin{figure}[t]
    \centering
    
    \includegraphics[width=0.48\columnwidth]{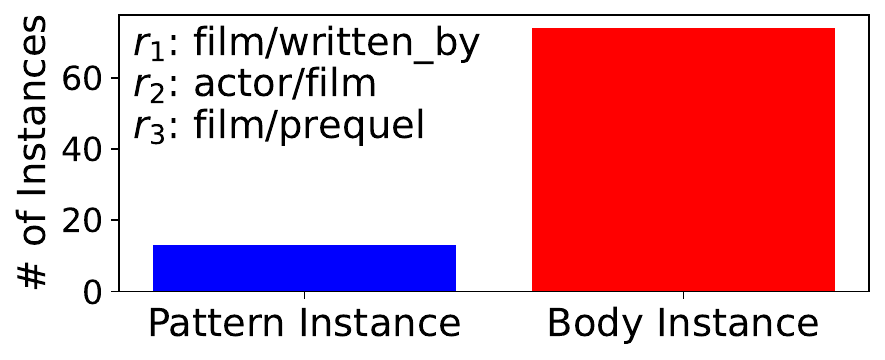}
    \label{fig:PI_BI_a}
    \hfill
    \includegraphics[width=0.48\columnwidth]{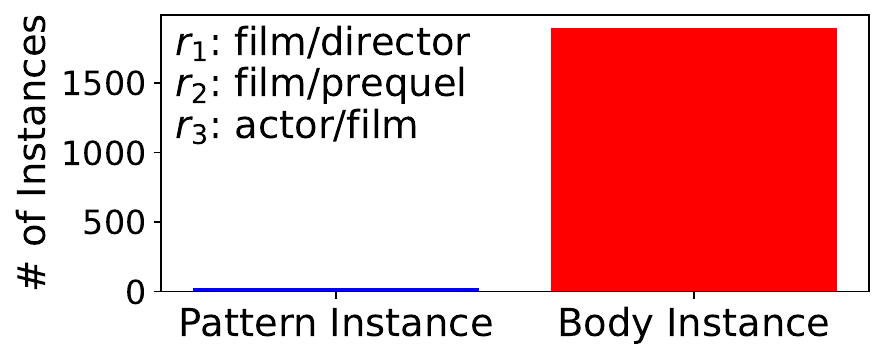}
    \label{fig:PI_BI_b}

    \caption{The number of pattern instances and body instances for the local patterns introduced in Figure~\ref{fig:problem a}.}
    \label{fig:PI_BI}
\end{figure}

\paragraph{Empirical Evidence}
Nevertheless, existing KGE models overlook the difference between local and universal patterns, treating all observed patterns as universally valid regardless of their frequency. Figure~\ref{fig:problem} shows the histograms of the embedding difference $\Delta = r_1^{T} \circ r_2^{T} \circ r_3^{H} - r_1^{H} \circ r_2^{H} \circ r_3^{T}$, that is presented in Equation~\ref{eq:pattern equation}. Elements of $\Delta$ close to zero indicate that the model recognizes the given relation set $(r_1, r_2, r_3)$ as a valid pattern, therefore, the model generalizes the pattern universally to every body instance. Figures~\ref{fig:problem a} and ~\ref{fig:problem b} show that the elements of $\Delta$ are concentrated near zero for both local and universal patterns, indicating that the model recognizes both as valid composition patterns regardless of instance frequency. This empirically demonstrates that PairRE is trained to generalize local patterns as if they were universal, even when the supporting instances are scarce. 

While local patterns are supported by only a scarce number of pattern instances, they often have a vast number of body instances. Figure~\ref{fig:PI_BI} presents the number of pattern instances and body instances of the local patterns in Figure~\ref{fig:problem a}. This indicates that a large number of body instances are affected by only a few pattern instances, leading the model to predict the corresponding head instances as valid for all body instances. The distribution of pattern instances and body instances of the universal patterns in Figure~\ref{fig:problem b} and empirical evidence for another pattern type are presented in Appendix~\ref{appendix:PI_BI_comparsion}.

To address this issue, we propose a novel KGE framework PogRE that explicitly models the distinction between pattern universality and locality. Our core idea is to generalize patterns differentially based on their observation frequency in $G$, rather than generalizing all patterns equally.

\paragraph{Which Methods Are Affected?}
Table~\ref{tab:model_comparison} presents representative examples of KGE methods that suffer from over-generalization. To verify whether these models actually suffer from over-generalization, we propose the Over-Generalization (OG) ratio. Specifically, for the local patterns presented in Figure~\ref{fig:PI_BI}, we extract the head instances corresponding to the body instances and categorize them into True triples (if triples are in $G$) and False triples (others). The OG ratio is defined as the average score produced by a model for the True triples divided by the average score for the False triples. An OG ratio close to 1 indicates that a model suffers from over-generalization, as it assigns similar scores to both True and False triples. Conversely, an OG ratio closer to 0 implies that the model effectively avoids this issue by assigning higher scores to False triples than to True triples. PogRE exhibits a lower OG ratio than other models. This indicates that PogRE effectively addresses over-generalization. For more details of the OG ratio, please refer to Appendix~\ref{appendix:OG_Ratio}.

\begin{table}[t]
\tiny
\setlength{\tabcolsep}{4pt}
\centering
\begin{tabular}{lcccc}
\hline
\multirow{2}{*}{Model} & \multirow{2}{*}{Score Function} & Over-generalization & \multirow{2}{*}{OG ratio ($\downarrow$)} \\ \cline{3-3} 
                       &                                 & Sym/Asym/Inves/Comp & \\ \hline
TransE                 &  $\|h+r-t\|$                               &    -/\ding{51}/\ding{51}/\ding{51}        &     .927              \\
RotatE                 &  $\|h\circ r-t\|$                               &   \ding{51}/\ding{51}/\ding{51}/\ding{51}   &   .921   \\
PairRE                 &   $\|h\circ r^H-t\circ r^T\|$                              &    \ding{51}/\ding{51}/\ding{51}/\ding{51} &  .917   \\
CompoundE        &   $\|M_r\cdot h-\hat{M}_r \cdot t \|$                              &       \ding{51}/\ding{51}/\ding{51}/\ding{51}&   .910      \\ \hline
PogRE (Ours)                  &   $\|L_r  h_r- t_r\|$                              &   \ding{55}/\ding{55}/\ding{55}/\ding{55} &   \textbf{.869}   \\ \hline
\end{tabular}

\caption{Comparison between PogRE and KGE models. $h$ and $t$ denote head and tail embeddings and $h_r$ and $t_r$ indicate head and tail embeddings in the relation-specific space, as presented in Equation~\ref{eq:head_tail_definition}.}
\label{tab:model_comparison}
\end{table}

\section{Method}
In this section, we present the formulation of PogRE and provide a theoretical analysis demonstrating how PogRE addresses over-generalization.

\subsection{Pattern Over-Generalization Robust Embedding (PogRE)}

\paragraph{Final Form}
We define the score function as the distance between the head entity $h_r$ and tail entity $t_r$ in relation-specific space, after the linear transformation $L_r \in \mathbb{R}^{d\times d}$:
\begin{equation}
\textstyle f_r(h, t) = \| L_rh_r - t_r \|
\end{equation}
where $h_r, t_r \in \mathbb{R}^d$ denote the head and tail embeddings in relation-specific space, respectively.
\paragraph{Comparison Between Existing Linear Transformation Models}
Although existing models such as RESCAL~\cite{nickel2011three} and TransR~\cite{lin2015learning} employ dense linear transformations, they suffer from overfitting and representing relations as $\mathbb{R}^{n\times n}$ dense linear matrix incurs significant computational costs. As a result, recent KGE models rarely adopt such dense linear transformations. CompoundE~\cite{ge2023compounding} utilizes sparse affine operators; consequently, it suffers from over-generalization, as presented in Table~\ref{tab:model_comparison}. In contrast, PogRE addresses over-generalization by employing dense linear transformations through a QR decomposition-inspired method, which reduces computational costs. Detailed differences are presented in Appendix~\ref{appendix:difference_between_linear_models}.
\paragraph{QR Decomposition and Partial Sharing for Efficient Parameterization}

Linear transformation $L_r$ of PogRE is decomposed into a relation-specific orthogonal matrix $Q_r$ and an upper-triangular matrix $R$ that is shared across all relations:
\begin{equation}
\begin{aligned}
\textstyle L_r &= Q_r R \\
\textstyle Q_r &= H_1H_2...H_k
\end{aligned}
\end{equation}
In addition, $Q_r$ is approximated using a product of $k$ Householder reflections (where $k \ll d$). This approximation significantly reduces the number of parameters from $n_r d^2$ to $d(d+1)/2 + n_r k d$, where $n_r$ denotes the number of relations, thereby effectively reducing the model complexity. Details about computational complexity with respect to $k$ are presented in Appendix~\ref{sec:computational_complexity}.

Additionally, let $\bar{R}$ be the learnable upper-triangular parameter matrix. The final shared matrix $R$ is formulated as:
\begin{equation}
 R = \frac{\bar{R}}{\|\bar{R}\|_2}
\end{equation}
where $\|\cdot\|_2$ denotes the spectral norm. We argue that even for local patterns, the pattern should be generalized to entities that are not observed in the patterns but are semantically similar to entities that are observed in the patterns. Spectral Normalization (SN) enables this generalization by bounding the Lipschitz constant of the transformations to one~\cite{miyato2018spectral}. Detailed derivations are provided in the Appendix~\ref{appendix:constraint_matrices}.

\paragraph{Relation-Specific Affine Mapping for Expressive Power}
Sharing an upper-triangular matrix $R$ reduces the expressive power of relation-specific transformations. To address this limitation and enhance the model capacity, following \cite{ge2023compounding}, each entity is mapped into an relation-specific space via three affine operators before applying $L_r$. By employing homogeneous coordinates, these operators can be unified into a single matrix multiplication:
\begin{equation}
\begin{aligned}
\textstyle h_r &= M_r h, & t_r &= \hat{M}_r \cdot t \\
\textstyle M_r &= S_r \cdot R_r \cdot T_r, & \qquad \hat{M}_r &= \hat{S}_r \cdot \hat{R}_r \cdot \hat{T}_r
\end{aligned}
\label{eq:head_tail_definition}
\end{equation}
where $h, t$ are head and tail embeddings, $S_r, R_r$, and $T_r$ denote the scaling, rotation, and translation operators, and $\hat{S}_r, \hat{R}_r$, and $\hat{T}_r$ denote the scaling, rotation, and translation operators for tail entity embedding, respectively. This mapping strategy ensures that each relation has sufficient expressive power despite the shared components in $L_r$.

\paragraph{Optimization}
Following ~\citet{sun2019rotate}, we adopt self-adversarial negative sampling for training. The loss function can be written as:\begin{equation}
\begin{aligned}
\textstyle L &= -\log \sigma(\gamma-f_r(h, t)) \\
      &\textstyle \quad -\sum^n_{i=1}p(h_i', r, t_i')\log\sigma(f_r(h_i',t_i')-\gamma)
\end{aligned}
\label{eq:gamma}
\end{equation}
where $\sigma$ is the sigmoid function, $\gamma$ is a fixed margin, $(h_i', r, t_i')$ is the $i$-th negative triple and $p(h_i', r, t_i')$ is the weight of the negative triple, defined as:
\begin{equation}
 \textstyle p(h_j', r, t_j'|\{(h_i, r_i, t_i)\}) = \frac{\text{exp} \alpha f_r(h_j', t_j')}{\sum_i\text{exp} \alpha f_r(h_i', t_i')}
\label{eq:alpha}
\end{equation}

where $\alpha$ is the temperature of sampling.

\subsection{How Is Pattern Over-Generalization Addressed?}
\label{sec:theoretical_analysis}
To analyze how PogRE addresses over generalization, we first consider using only the linear transformation $L_r$, and then extend this analysis to our framework, which incorporates the affine operators.

\paragraph{Theoretical Analysis: Linear Transformation}

To the best of our knowledge, all patterns studied in existing research are based on connected paths formed by relations. This implies that a body ($\psi$) and head ($\phi$) can be represented as a relational path between the start entity $e_u$ and the end entity $e_v$, where each path is formulated as a product of linear matrix multiplications. Consequently, the body ($\psi$) and head ($\phi$) of a pattern can be expressed as:
\begin{equation}
\begin{aligned}
\textstyle \text{Body }(\psi): \quad & L_{\psi}e_u = L_{r_n} \dots L_{r_2} L_{r_1}e_u = e_v, \\
\textstyle  \text{Head }(\phi): \quad & L_{\phi}e_u = L_{r_m'} \dots L_{r_{2}'} L_{r_{1}'}e_u = e_v.
\end{aligned}
\label{eq:body_head_definition}
\end{equation}
where $L \in \mathbb{R}^{d \times d}$ and $e \in \mathbb{R}^{d}$ denote the transformation matrix and entity vector, respectively.

From Equation~\ref{eq:body_head_definition}, since both paths map $e_u$ to the same entity $e_v$, we explicitly have $L_{\psi}e_u = L_{\phi}e_u$, which is equivalent to:
\begin{equation}
\textstyle (L_{\psi} - L_{\phi})e_u = 0 \quad \iff \quad E e_u = 0
\end{equation}
where $E = L_{\psi} - L_{\phi}$ denotes the constraint matrix.
Next, consider a set of $d$ linearly independent entities $\{e_{u_1}, e_{u_2}, \dots, e_{u_{d}}\}$ that satisfy the pattern, such that\footnote{Appendix~\ref{appendix:constraint_matrices} details constraint matrix for various patterns.}:
\begin{equation}
\textstyle E e_{u_i} = 0, \quad \text{for all } i = 1, 2, \dots, d
\end{equation}
For any arbitrary entity $a \in \mathbb{R}^{d}$, since $\{e_{u_i}\}$ forms a basis in $\mathbb{R}^{d}$, $a$ can be expressed as a linear combination $a = c_1e_{u_1} + c_2e_{u_2} + \dots + c_{d}e_{u_{d}}$. By the linearity of the transformation $E$, it follows that:
\begin{equation}
\textstyle E a = c_1 E e_{u_1} + c_2 E e_{u_2} + \dots + c_{d} E e_{u_{d}} = 0
\end{equation}

These results indicate that as PogRE observes more linearly independent entities $e_u$, the dimension of the space spanned by these entities increases. Consequently, when the number of observed entities reaches $d$, PogRE guarantees the universal generalization of the pattern across all related instances. In other words, it can be expected that a pattern becomes progressively universal as the number of observed entities increases.

\paragraph{Extension to Relation-Specific Affine Mapping}

This analysis can be extended to our proposed framework by employing homogeneous coordinates. By representing entities in an augmented $(d+1)$-dimensional space, the integration of affine operators and linear transformations for a relation $r$ can be unified into a single linear matrix $A_r \in \mathbb{R}^{(d+1) \times (d+1)}$ when $\hat{M}_r$ is non-singular:
\begin{equation}
\textstyle A_r = \hat{M}_r^{-1} L_r M_r
\end{equation}
Therefore, the relational path can be expressed as a product of linear transformation $A_r$. Consequently, the same proof used in the linear case can be applied, demonstrating that the pattern becomes universal only when $d+1$ linearly independent entities are observed in the augmented space. The linear independence of entity embeddings is discussed in Section~\ref{sec:experiment_independent}. Our theoretical guarantees rely on the ideal assumption that $\|Ee_i\|=0$. Since satisfying this exact constraint is challenging in practice, we provide further analysis in Appendix~\ref{appendix:Bounding_constraint_error}, proving that an approximate constraint ($\|Ee_i\| < \epsilon$) still bounds the pattern constraint of unseen entities, along with a discussion on the practical strength of the approximate constraint assumption.

\section{Related Work}
\label{sec:related_work}
\subsection{Distance-based Models}
Distance-based models capture patterns through various relational operations. TransE~\cite{bordes2013translating}, RotatE~\cite{sun2019rotate}, Rotate3D~\cite{gao2020rotate3d}, DualE~\cite{cao2021dual}, ReflectE~\cite{zhang2022knowledge} and, RotatQ~\cite{xie2025rotatq} model relations through translation, rotation, 3D rotation, a combination of translation and rotation, reflection transformation, and quaternion-based transformation, respectively. Other models enrich these operations: HAKE~\cite{zhang2020learning} uses polar coordinates for semantic hierarchies, PairRE~\cite{chao2021pairre} and CompoundE~\cite{ge2023compounding} apply scaling and compound operators, and DensE~\cite{lu2022dense} decomposes relations into rotation and scaling in 3D Euclidean space. Recent models further diversify relation modeling: ExpressivE~\cite{pavlovicexpressive} and OctagonE~\cite{charpenay2024capturing} represent relations as hyper-parallelograms and axis-aligned octagons, respectively. SpeedE~\cite{pavlovic2024speede} improves efficiency in low-dimensional Euclidean settings, OrthogonalE~\cite{zhu2024block} adopts block-diagonal orthogonal matrices with Riemannian optimization, and \citet{charpenay2025less} theoretically analyze the ability of MuRE to capture inference patterns. Although these models effectively capture patterns, their pattern conditions are determined by relation embeddings, which can generalize patterns even when the pattern is supported by only a few instances.

\begin{table*}[t]
\centering
\scriptsize
\renewcommand{\arraystretch}{1}
\begin{tabular}{lccccccccc}
\hline
\multirow{2}{*}{Knowledge Graph Embedding} & \multicolumn{3}{c}{WN18RR} & \multicolumn{3}{c}{FB15k-237} & \multicolumn{3}{c}{YAGO3-10} \\ \cline{2-10} 
                                           & MRR     & H@1     & H@10   & MRR      & H@1      & H@10    & MRR      & H@1     & H@10    \\ \hline
TransE~\cite{bordes2013translating}                                     & .226    & -       & .501   & .294     & -        & .465    & -        & -       & -       \\
DistMult~\cite{yang2015embedding}                                   & .430    & .390    & .490   & .241     & .155     & .419    & -        & -       & -       \\
ComplEx~\cite{trouillon2016complex}                                    & .440    & .410    & .510   & .247     & .158     & .428    & -        & -       & -       \\
RotatE~\cite{sun2019rotate}                                     & .476    & .428    & .571   & .338     & .241     & .533    & .495     & .402    & .670    \\
TuckER~\cite{balavzevic2019tucker}                                     & .470    & .443    & .526   & .358     & .266     & .544    & -        & -       & -       \\
QuatE~\cite{zhang2019quaternion}                                      & .488    & .438    & .582   & .348     & .248     & .550    & -        & -       & -       \\
Rotate3D~\cite{gao2020rotate3d}                                       & .489    & .442    & .579   & .347     & .250     & .543    & -     & -    & -    \\
HAKE~\cite{zhang2020learning}                                       & \underline{.497}    & \underline{.452}    & .582   & .346     & .250     & .542    & \underline{.545}     & \underline{.462}    & .694    \\
DualE~\cite{cao2021dual}                                      & .492    & .444    & .584   & \underline{.365}     & \underline{.268}     & .559    & -        & -       & -       \\
PairRE~\cite{chao2021pairre}                                     & -       & -       & -      & .351     & .256     & .544    & -        & -       & -       \\
HopfE~\cite{bastos2021hopfe}                                      & .472    & .413    & .586   & .343     & .247     & .534    & .529     & .438    & \underline{.695}    \\
DensE~\cite{lu2022dense}                                      & .492    & -       & .586   & .351     & -        & .544    & .541     & -       & .678    \\
ReflectE~\cite{zhang2022knowledge}                                   & .488    & .450    & .559   & .358     & .263     & .546    & -        & -       & -       \\
ExpressivE~\cite{pavlovicexpressive}                                 & .482    & .407    & \textbf{.619}   & .350     & .256     & .535    & -        & -       & -       \\
CompoundE~\cite{ge2023compounding}                                  & .491    & .450    & .576   & .357     & .264     & .545    & -        & -       & -       \\
SpeedE~\cite{pavlovic2024speede}                                     & .493    & .446    & -      & .320     & .227     & -       & .413     & .332    & -       \\
OctagonE~\cite{charpenay2024capturing}                          & .479    & .436    & .561   & .332     & .241     & .517    & -        & -       & -       \\
OrthogonalE~\cite{zhu2024block}                                & .494    & .446    & .573   & .334     & .242     & .518    & -        & -       & -       \\ 
CustomizE~\cite{guan2025should}  & .486    & .446    & -   & .351     & .261     & .504    & -        & -       & -       \\
RotatQ~\cite{xie2025rotatq}  & .489    & .450    & .552   & .356     & .254     & \textbf{.619}    & -        & -       & -       \\
MuRE variant~\cite{charpenay2025less} & .469    & .427    & .553   & .307     & .212     & .503    & -        & -       & -       \\

\hline
\multirow{2}{*}{PogRE (Ours)}                      &   \textbf{.506}      & \textbf{.461}        &  \underline{.595}      & \textbf{.369}     & \textbf{.273}     & \underline{.562}    & \textbf{.556}     & \textbf{.474}    & \textbf{.699}    \\
                                          &       $\pm$.001  &      $\pm$.001   &  $\pm$.000     & $\pm$.001   & $\pm$.001   & $\pm$.001  &    $\pm$.000      &     $\pm$.001    &     $\pm$.000    \\ \hline
\end{tabular}

\caption{Link prediction results on WN18RR, FB15k-237 and YAGO3-10. Bold indicates the best result and underline indicates the second best result. $\pm$ indicates standard deviation.} 
\label{tab:main_results}
\end{table*}

\subsection{Tensor Decomposition Models}
Tensor decomposition models capture patterns through interactions among entity and relation embeddings. DistMult~\cite{yang2015embedding} and ComplEx~\cite{trouillon2016complex} use bilinear scoring functions, whereas HolE~\cite{nickel2016holographic} employs circular correlation. ANALOGY~\cite{liu2017analogical}, SimplE~\cite{kazemi2018simple}, and TuckER~\cite{balavzevic2019tucker} use normal linear operators, enhanced CP decomposition, and Tucker decomposition, respectively. QuatE~\cite{zhang2019quaternion} extends interactions with quaternion representations, while CustomizE~\cite{guan2025should} introduces customized embeddings to address the long-tail problem. Although these models provide strong representation capacity, they are not explicitly designed to distinguish between local patterns and universal patterns. As a result, they may capture observed patterns, but they do not directly control the scope of pattern generalization based on supporting evidence.

\section{Experiments}
\subsection{Experimental Setting}

\paragraph{Dataset}
We evaluate PogRE on three widely used KG datasets: WN18RR~\cite{dettmers2018convolutional}, FB15k-237~\cite{toutanova2015observed} and YAGO3-10~\cite{mahdisoltani2013yago3}. The statistics of these datasets are presented in Appendix~\ref{appendix:datasets}

\paragraph{Evaluation Protocol}
We evaluate link prediction performance in the filtered setting~\cite{bordes2013translating}. In this setting, test triples are ranked against all other candidate triples that are generated by corrupting subjects or objects: ${(h', r, t)}$ or ${(h, r, t')}$, and all the triples that appear either in the training, validation or test set are removed from the candidate triples, except the test triple of interest. We adopt MRR, Hits@1 (H@1), and Hits@10 (H@10) to compare the performance of different KGE models. MRR denotes the mean reciprocal rank of the correct entities, and H@N represents the proportion of correct entities ranked within the top $N$. For performance comparison, we evaluate PogRE against all KGE models discussed in Section~\ref{sec:related_work}.

\begin{table}[t]
\scriptsize
\centering
\begin{tabular}{lccc}
\hline
\multirow{2}{*}{Model} & \multicolumn{3}{c}{MRR}                                                                   \\ \cline{2-4} 
                       & \multicolumn{1}{l}{WN18RR} & \multicolumn{1}{l}{FB15k-237} & \multicolumn{1}{l}{YAGO3-10} \\ \hline
PogRE                   & .506                       & .369                          & .556                         \\ \hline
+ w/o $R$                & .505                       & .365                          & .517                         \\
+ w/o $Q_{r}$                & .501                       & .364                          & .540                         \\
+ w/o $SN$                & .505                       & .360                          & .523       
 \\ 
 + w/o $L_r$ (CompoundE)              & .491                       & .357                          & .477  \\ 
  + w/o QR Decomposition              & OOM                       & OOM                          & OOM   
 \\
 \hline
\end{tabular}

\caption{Ablation study of PogRE on WN18RR, FB15k-237 and YAGO3-10. MRR is used for performance comparison. $R$, $Q_r$, $SN$, $L_r$, and OOM denote the shared upper triangular parameter matrix, relation specific Householder reflection, Spectral Normalization, linear transformation of PogRE, and Out of Memory, respectively. In w/o QR Decomposition, $n\times n$ dense linear transformations are used for $L_r$.}
\label{tab:ablation_study}
\end{table}

\subsection{Main Results}
\paragraph{Link Prediction Performance}
As shown in Table \ref{tab:main_results}, PogRE exhibits superior or competitive performance compared with the baselines. For instance, PogRE achieves MRR improvements of 0.009, 0.004, and 0.011 over the second-best models, DualE and HAKE, on WN18RR, FB15k-237 and YAGO3-10, respectively. These results indicate the effectiveness and robustness of PogRE across diverse datasets. In addition to the standard benchmarks presented above, Appendix~\ref{appendix:large-scale_KG} provides link prediction results on large-scale KG datasets.

\paragraph{Ablation Study}
Table~\ref{tab:ablation_study} summarizes the results of an ablation study conducted to verify the effectiveness of each proposed component. As shown in the results, PogRE consistently outperforms the ablated models across all datasets. Specifically, w/o $L_r$ (equivalent to CompoundE) exhibits significant performance degradation. w/o $L_r$ does not employ a dense matrix and thus suffers from over-generalization, which suggests that overlooking this problem results in significant performance loss. Furthermore, employing dense linear transformations without QR decomposition was infeasible across all datasets; this demonstrates that models such as TransR~\cite{lin2015learning}, which rely on dense linear transformations, lack scalability due to their high computational costs.

\subsection{Analysis}
\paragraph{Entity Independence}
\label{sec:experiment_independent}

As discussed in Section~\ref{sec:theoretical_analysis}, PogRE ensures that any pattern becomes progressively universal as more linearly independent entities are observed in the pattern. This implies that if the entity embeddings trained by PogRE are linearly independent, PogRE can achieve such progressive universality in practice. To investigate entity independence, we randomly sample entity embeddings trained by PogRE and compute the rank of the space spanned by the sampled entities. Table~\ref{tab:space_span} shows the mean rank of the subspace spanned by sampled entities over 100 random trials. We empirically observe that the rank of the space spanned by the randomly sampled entities is approximately equal to the number of sampled entities, demonstrating that the sampled entities are linearly independent. These results indicate that, since entities in practice are shown to be linearly independent, the dimension of the space spanned by the entities increases as the number of observed entities increases, and that universal generalization is achieved when around $d+1$ entities are observed. We also present empirical results on the independence of entities observed in specific patterns in Appendix~\ref{appendix:linear_independence_within_pattern}.

\begin{table}[t]
\scriptsize
\centering
\begin{tabular}{lccc}
\hline
\multirow{2}{*}{\begin{tabular}[l]{@{}l@{}}Number of \\ Sample Vector\end{tabular}} & \multicolumn{3}{c}{Rank}                                                                   \\ \cline{2-4} 
                       & \multicolumn{1}{l}{WN18RR} & \multicolumn{1}{l}{FB15k-237} & \multicolumn{1}{l}{YAGO3-10} \\ \hline 
100    & 100.0       & 100.0                      & 100.0                                                 \\
200    & 200.0       & 200.0                      & 200.0                                                 \\
500    & 500.0       & 500.0                      & 500.0                                                 \\
1,000  & 994.4 $\pm$ 1.5      & 1000.0                     & 999.5 $\pm$ 0.5                                               \\
1,500  & 1000.0       & 1497.4 $\pm$ 0.6                   & 1000.0                                               \\
2,000  & 1000.0       & 1500.0                     & 1000.0                                               \\ \hline 
\end{tabular}

\caption{Mean rank of the subspace spanned by randomly sampled entities over 100 random trials across three benchmarks. The entity dimensions of PogRE are 1,000 on WN18RR and YAGO3-10, and 1,500 on FB15k-237.}
\label{tab:space_span}
\end{table}

\begin{table}[t]
\centering
\resizebox{\columnwidth}{!}{%
\begin{tabular}{ccccccc}
\hline

\multirow{2}{*}{\begin{tabular}[c]{@{}c@{}}\# of Pattern \\ Instances (n)\end{tabular}} & \multicolumn{2}{c}{WN18RR}                                          & \multicolumn{2}{c}{FB15k-237}                                      & \multicolumn{2}{c}{YAGO3-10}                                       \\ \cline{2-7}    & \multicolumn{1}{l}{\# of Patterns} & \multicolumn{1}{l}{Prop. (\%)} & \multicolumn{1}{l}{\# of Patterns} & \multicolumn{1}{l}{Prop. (\%)} & \multicolumn{1}{l}{\# of Patterns} & \multicolumn{1}{l}{Prop. (\%)} \\ \hline
n = 1                                                                                   & 17                              & 48.6                           & 1,546                           & 26.3                           & 56                              & 17.6                           \\
1 $<$ n $\leq$ 10                                                                       & 14                              & 40.0                           & 2,288                           & 39.0                           & 112                             & 35.2                           \\
10 $<$ n $\leq 10^2$                                                                    & 4                               & 11.4                           & 1,439                           & 24.5                           & 91                              & 28.6                           \\
$10^2 <$ n $\leq 10^3$                                                                  & -                               & -                              & 489                             & 8.3                            & 53                              & 16.7                           \\
n $> 10^3$                                                                              & -                               & -                              & 111                             & 1.9                            & 6                               & 1.9                            \\ \hline
Total                                                                                   & 35                              & 100\%                          & 5,873                           & 100\%                          & 318                             & 100\%                          \\ \hline
\end{tabular}}

\caption{Distribution of composition patterns according to the number of pattern instances ($n$) across three benchmark datasets. \# of Pattern Instances and \# of Patterns indicate the number of pattern instances and the number of patterns, respectively. Prop. (\%) is calculated as the number of patterns within each range of $n$ divided by the total number of patterns, within each dataset.}

\label{tab:composition_pattern_number}
\end{table}
\paragraph{Distribution of Patterns by Number of Pattern Instances}
Table~\ref{tab:composition_pattern_number} presents the distribution of composition patterns according to the number of their pattern instances $n$ across three KG benchmarks. We compute the number of composition patterns as the number of relation sets $(r_x, r_y, r_z) \in R$ that have at least one observed composition pattern instance $(\psi_1 \Rightarrow \phi_1)$ in the KG. In WN18RR, FB15k-237 and YAGO3-10, 88.6\%, 65.3\% and 52.8\% of composition patterns have 10 or fewer pattern instances, respectively. This result indicates that a substantial proportion of patterns in KGs are observed in only a few instances. These patterns can be generalized universally when a model suffers from over-generalization. The distributions of other patterns are presented in Appendix~\ref{appendix:over-generalization_experiments}.

\begin{figure}[t]
    \centering
    \includegraphics[width=1\columnwidth]{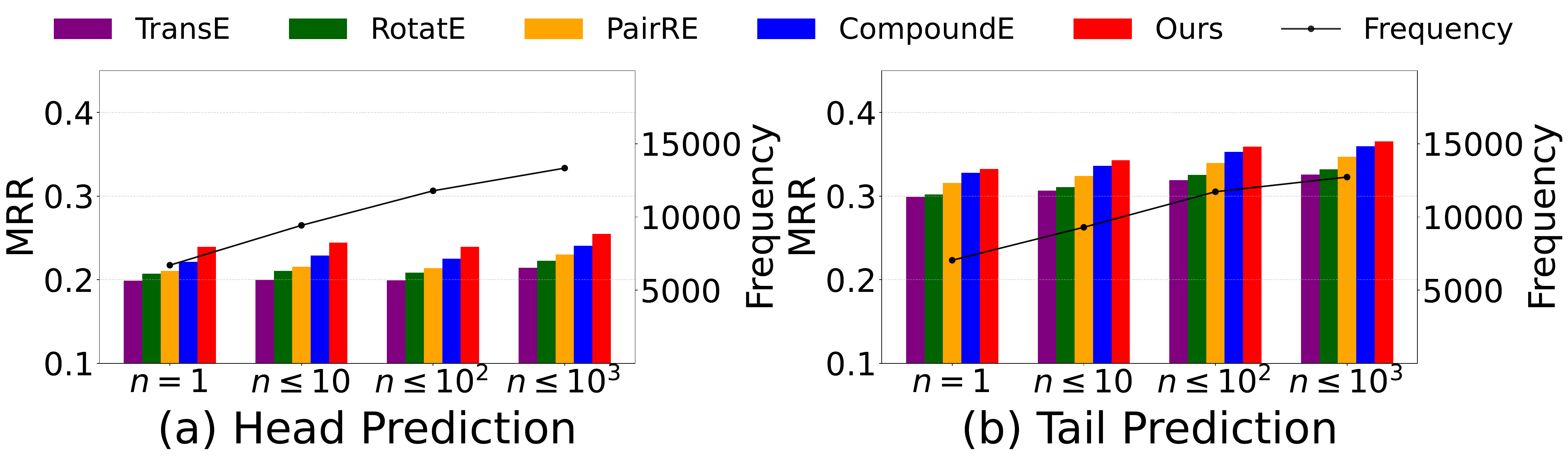}
    \caption{MRR comparison between PogRE and baseline models for various sparsity conditions of pattern instances on FB15k-237. The black line indicates the number of test triples of $G_{over}$.}
    \label{fig:composition_experiments}
\end{figure}

\paragraph{Quantified Impact of Over-generalization}
To investigate the impact of over-generalization, we extract $G_{over}$, a set of triples $(h, r, t)$, where candidates $(h, r, t')$ or $(h', r, t)$ (with $t' \neq t$ and $h' \neq h$) are heads of pattern instances whose bodies are in the training set, i.e., candidates $(h, r, t')$ or $(h', r, t)$ are $\phi_i \in G_u \setminus G_o$ for which there is a corresponding $\psi_i \in G_o$. Intuitively, if the model suffers from over-generalization, the rank of $(h, r, t)$ is lower than the rank of the candidate triples. To extract $G_{over}$, we consider symmetry, inversion, composition, hierarchy, intersection, transitive, g.intersection, b. transitive and b. composition where the number of pattern instances is $n =1$, $n \leq 10$, $n \leq 10^2$, and $n \leq 10^3$. We compare the MRR of PogRE with other baselines: TransE, RotatE, PairRE, and CompoundE on $G_{over}$. Figure~\ref{fig:composition_experiments} presents the results on FB15k-237. We observe that PogRE consistently outperforms the baselines, regardless of the number of pattern instances. These results show that PogRE effectively addresses the negative impact of over-generalization. The detailed definitions of $G_{over}$ and the comparison results for WN18RR and YAGO3-10 are presented in Appendix~\ref{appendix:over-generalization_experiments}.

\section{Conclusion}
In this paper, we propose PogRE, a novel KGE method that utilizes linear transformations and compound operations. PogRE addresses over-generalization, a phenomenon in which a model generalizes a pattern to every body instance in the graph after observing only a single instance. Our theoretical analysis shows that PogRE allows a pattern to become progressively universal as more linearly independent entities are observed. Experimental results on three benchmark datasets demonstrate the effectiveness of PogRE.
\section*{Limitations}
To universally generalize patterns, PogRE does not utilize the semantics of patterns, which can be a useful inductive bias for pattern generalization. Therefore, for universal but low-frequency patterns, PogRE may fail to generalize them universally, and for local but high-frequency patterns, PogRE may generalize them universally, resulting in inappropriate generalization. This limitation arises when pattern frequency does not align with semantic universality. Accordingly, PogRE should be understood as alleviating, rather than fully resolving, pattern over-generalization. To address this limitation, in future work, we will leverage the semantics of patterns for pattern generalization.

Furthermore, PogRE is limited to the transductive setting, where the goal is to learn and improve embedding structures for a fixed set of known entities and relations. Since PogRE explicitly learns entity and relation embeddings for entities and relations observed during training, it cannot directly represent entities or relations not observed during training. While extending PogRE to the inductive setting is an important problem for handling unknown entities and relations, addressing it requires substantially different assumptions and architectural designs. For this reason, we leave extending PogRE to the inductive setting as future work.

\section*{Acknowledgements}
This work was supported by the National Research Foundation of Korea(NRF) grant funded by the Korea government(MSIT) (RS-2026-25520248) (Contribution Rate: 50\%);  the National Research Foundation of Korea (NRF) grant funded by the Korea government (MSIT) (No.2022R1A2C2012054, Development of AI for Canonicalized Expression of Trained Hypotheses by Resolving Ambiguity in Various Relation Levels of Representation Learning) (Contribution Rate: 40\%); and Institute of Information \& communications Technology Planning \& Evaluation (IITP) grant funded by the Korea government (MSIT) (No.2019-0-01842, Artificial Intelligence Graduate School Program (GIST)) (Contribution Rate: 10\%).

\bibliography{custom}

@article{bordes2013translating,
  title={Translating embeddings for modeling multi-relational data},
  author={Bordes, Antoine and Usunier, Nicolas and Garcia-Duran, Alberto and Weston, Jason and Yakhnenko, Oksana},
  journal={Advances in neural information processing systems},
  volume={26},
  year={2013}
}

@inproceedings{yang2015embedding,
  title={Embedding Entities and Relations for Learning and Inference in Knowledge Bases},
  author={Yang, Bishan and Yih, Scott Wen-tau and He, Xiaodong and Gao, Jianfeng and Deng, Li},
  booktitle={Proceedings of the International Conference on Learning Representations (ICLR) 2015},
  year={2015}
}

@inproceedings{trouillon2016complex,
  title={Complex embeddings for simple link prediction},
  author={Trouillon, Th{\'e}o and Welbl, Johannes and Riedel, Sebastian and Gaussier, {\'E}ric and Bouchard, Guillaume},
  booktitle={International conference on machine learning},
  pages={2071--2080},
  year={2016},
  organization={PMLR}
}

@inproceedings{sun2019rotate,
  title={RotatE: Knowledge Graph Embedding by Relational Rotation in Complex Space},
  author={Sun, Zhiqing and Deng, Zhi-Hong and Nie, Jian-Yun and Tang, Jian},
  booktitle={International Conference on Learning Representations},
  year={2019}
}

@inproceedings{balavzevic2019tucker,
  title={TuckER: Tensor Factorization for Knowledge Graph Completion},
  author={Bala{\v{z}}evi{\'c}, Ivana and Allen, Carl and Hospedales, Timothy},
  booktitle={Proceedings of the 2019 Conference on Empirical Methods in Natural Language Processing and the 9th International Joint Conference on Natural Language Processing (EMNLP-IJCNLP)},
  pages={5185--5194},
  year={2019}
}

@article{zhang2019quaternion,
  title={Quaternion knowledge graph embeddings},
  author={Zhang, Shuai and Tay, Yi and Yao, Lina and Liu, Qi},
  journal={Advances in neural information processing systems},
  volume={32},
  year={2019}
}

@inproceedings{zhang2020learning,
  title={Learning hierarchy-aware knowledge graph embeddings for link prediction},
  author={Zhang, Zhanqiu and Cai, Jianyu and Zhang, Yongdong and Wang, Jie},
  booktitle={Proceedings of the AAAI conference on artificial intelligence},
  volume={34},
  pages={3065--3072},
  year={2020}
}

@inproceedings{cao2021dual,
  title={Dual quaternion knowledge graph embeddings},
  author={Cao, Zongsheng and Xu, Qianqian and Yang, Zhiyong and Cao, Xiaochun and Huang, Qingming},
  booktitle={Proceedings of the AAAI conference on artificial intelligence},
  volume={35},
  pages={6894--6902},
  year={2021}
}

@inproceedings{chao2021pairre,
  title={PairRE: Knowledge Graph Embeddings via Paired Relation Vectors},
  author={Chao, Linlin and He, Jianshan and Wang, Taifeng and Chu, Wei},
  booktitle={Proceedings of the 59th Annual Meeting of the Association for Computational Linguistics and the 11th International Joint Conference on Natural Language Processing (Volume 1: Long Papers)},
  pages={4360--4369},
  year={2021}
}

@inproceedings{bastos2021hopfe,
  title={Hopfe: Knowledge graph representation learning using inverse hopf fibrations},
  author={Bastos, Anson and Singh, Kuldeep and Nadgeri, Abhishek and Shekarpour, Saeedeh and Mulang, Isaiah Onando and Hoffart, Johannes},
  booktitle={Proceedings of the 30th ACM international conference on information \& knowledge management},
  pages={89--99},
  year={2021}
}

@article{lu2022dense,
  title={DensE: An enhanced non-commutative representation for knowledge graph embedding with adaptive semantic hierarchy},
  author={Lu, Haonan and Hu, Hailin and Lin, Xiaodong},
  journal={Neurocomputing},
  volume={476},
  pages={115--125},
  year={2022},
  publisher={Elsevier}
}

@article{zhang2022knowledge,
  title={Knowledge graph embedding by reflection transformation},
  author={Zhang, Qianjin and Wang, Ronggui and Yang, Juan and Xue, Lixia},
  journal={Knowledge-Based Systems},
  volume={238},
  pages={107861},
  year={2022},
  publisher={Elsevier}
}

@inproceedings{pavlovicexpressive,
  title={ExpressivE: A Spatio-Functional Embedding For Knowledge Graph Completion},
  author={Pavlovi{\'c}, Aleksandar and Sallinger, Emanuel},
  booktitle={The Eleventh International Conference on Learning Representations},
  year={2023}
}

@inproceedings{ge2023compounding,
  title={Compounding geometric operations for knowledge graph completion},
  author={Ge, Xiou and Wang, Yun Cheng and Wang, Bin and Kuo, C-C Jay},
  booktitle={Proceedings of the 61st Annual Meeting of the Association for Computational Linguistics (Volume 1: Long Papers)},
  pages={6947--6965},
  year={2023}
}

@inproceedings{pavlovic2024speede,
  title={SpeedE: Euclidean Geometric Knowledge Graph Embedding Strikes Back},
  author={Pavlovi{\'c}, Aleksandar and Sallinger, Emanuel},
  booktitle={Findings of the Association for Computational Linguistics: NAACL 2024},
  pages={69--92},
  year={2024}
}

@inproceedings{charpenay2024capturing,
  title={Capturing knowledge graphs and rules with octagon embeddings},
  author={Charpenay, Victor and Schockaert, Steven},
  booktitle={Proceedings of the Thirty-Third International Joint Conference on Artificial Intelligence},
  pages={3289--3297},
  year={2024}
}

@inproceedings{zhu2024block,
  title={Block-Diagonal Orthogonal Relation and Matrix Entity for Knowledge Graph Embedding},
  author={Zhu, Yihua and Shimodaira, Hidetoshi},
  booktitle={Findings of the Association for Computational Linguistics: EMNLP 2024},
  pages={16956--16972},
  year={2024}
}

@inproceedings{gao2020rotate3d,
  title={Rotate3d: Representing relations as rotations in three-dimensional space for knowledge graph embedding},
  author={Gao, Chang and Sun, Chengjie and Shan, Lili and Lin, Lei and Wang, Mingjiang},
  booktitle={Proceedings of the 29th ACM international conference on information \& knowledge management},
  pages={385--394},
  year={2020}
}

@inproceedings{krishnan2024method,
  title={A method for assessing inference patterns captured by embedding models in knowledge graphs},
  author={Krishnan, Narayanan Asuri and Rivero, Carlos R},
  booktitle={Proceedings of the ACM Web Conference 2024},
  pages={2030--2041},
  year={2024}
}

@inproceedings{miyato2018spectral,
  title={Spectral Normalization for Generative Adversarial Networks},
  author={Miyato, Takeru and Kataoka, Toshiki and Koyama, Masanori and Yoshida, Yuichi},
  booktitle={International Conference on Learning Representations},
  year={2018}
}

@inproceedings{dettmers2018convolutional,
  title={Convolutional 2d knowledge graph embeddings},
  author={Dettmers, Tim and Minervini, Pasquale and Stenetorp, Pontus and Riedel, Sebastian},
  booktitle={Proceedings of the AAAI conference on artificial intelligence},
  volume={32},
  year={2018}
}

@inproceedings{toutanova2015observed,
  title={Observed versus latent features for knowledge base and text inference},
  author={Toutanova, Kristina and Chen, Danqi},
  booktitle={Proceedings of the 3rd workshop on continuous vector space models and their compositionality},
  pages={57--66},
  year={2015}
}

@inproceedings{mahdisoltani2013yago3,
  title={Yago3: A knowledge base from multilingual wikipedias},
  author={Mahdisoltani, Farzaneh and Biega, Joanna and Suchanek, Fabian M},
  booktitle={CIDR},
  year={2013}
}

@article{kazemi2018simple,
  title={Simple embedding for link prediction in knowledge graphs},
  author={Kazemi, Seyed Mehran and Poole, David},
  journal={Advances in neural information processing systems},
  volume={31},
  year={2018}
}

@inproceedings{liu2017analogical,
  title={Analogical inference for multi-relational embeddings},
  author={Liu, Hanxiao and Wu, Yuexin and Yang, Yiming},
  booktitle={International conference on machine learning},
  pages={2168--2178},
  year={2017},
  organization={PMLR}
}

@inproceedings{nickel2016holographic,
  title={Holographic embeddings of knowledge graphs},
  author={Nickel, Maximilian and Rosasco, Lorenzo and Poggio, Tomaso},
  booktitle={Proceedings of the AAAI conference on artificial intelligence},
  volume={30},
  number={1},
  year={2016}
}

@inproceedings{bollacker2008freebase,
  title={Freebase: a collaboratively created graph database for structuring human knowledge},
  author={Bollacker, Kurt and Evans, Colin and Paritosh, Praveen and Sturge, Tim and Taylor, Jamie},
  booktitle={Proceedings of the 2008 ACM SIGMOD international conference on Management of data},
  pages={1247--1250},
  year={2008}
}

@article{miller1995wordnet,
  title={WordNet: a lexical database for English},
  author={Miller, George A},
  journal={Communications of the ACM},
  volume={38},
  number={11},
  pages={39--41},
  year={1995},
  publisher={ACM New York, NY, USA}
}

@inproceedings{ma-etal-2025-large-language-models-meet,
    title = "Large Language Models Meet Knowledge Graphs for Question Answering: Synthesis and Opportunities",
    author = "Ma, Chuangtao  and
      Chen, Yongrui  and
      Wu, Tianxing  and
      Khan, Arijit  and
      Wang, Haofen",
    editor = "Christodoulopoulos, Christos  and
      Chakraborty, Tanmoy  and
      Rose, Carolyn  and
      Peng, Violet",
    booktitle = "Proceedings of the 2025 Conference on Empirical Methods in Natural Language Processing",
    month = nov,
    year = "2025",
    address = "Suzhou, China",
    publisher = "Association for Computational Linguistics",
    url = "https://aclanthology.org/2025.emnlp-main.1249/",
    doi = "10.18653/v1/2025.emnlp-main.1249",
    pages = "24589--24608",
    ISBN = "979-8-89176-332-6"
}

@inproceedings{sui2025fidelis,
  title={Fidelis: Faithful reasoning in large language models for knowledge graph question answering},
  author={Sui, Yuan and He, Yufei and Liu, Nian and He, Xiaoxin and Wang, Kun and Hooi, Bryan},
  booktitle={Findings of the Association for Computational Linguistics: ACL 2025},
  pages={8315--8330},
  year={2025}
}

@inproceedings{nickel2011three,
  title={A three-way model for collective learning on multi-relational data},
  author={Nickel, Maximilian and Tresp, Volker and Kriegel, Hans-Peter},
  booktitle={Proceedings of the 28th International Conference on International Conference on Machine Learning},
  pages={809--816},
  year={2011}
}

@inproceedings{lin2015learning,
  title={Learning entity and relation embeddings for knowledge graph completion},
  author={Lin, Yankai and Liu, Zhiyuan and Sun, Maosong and Liu, Yang and Zhu, Xuan},
  booktitle={Proceedings of the AAAI conference on artificial intelligence},
  volume={29},
  number={1},
  year={2015}
}

@article{xie2025rotatq,
  title={RotatQ: Knowledge graph embedding based on quaternion unit},
  author={Xie, Shiwen and Xie, Yongfang and Hu, Cheng and Huang, Tingwen},
  journal={Neurocomputing},
  pages={132413},
  year={2025},
  publisher={Elsevier}
}

@inproceedings{guan2025should,
  title={Should We Use a Fixed Embedding Size? Customized Dimension Sizes for Knowledge Graph Embedding},
  author={Guan, Zhanpeng and Zhang, Zhao and Wu, Yiqing and Zhang, Fuwei and Xu, Yongjun},
  booktitle={Proceedings of the 31st International Conference on Computational Linguistics},
  pages={9006--9012},
  year={2025}
}

@inproceedings{charpenay2025less,
  title={Less Is MuRE: Revisiting shallow knowledge graph embeddings},
  author={Charpenay, Victor and Schockaert, Steven},
  booktitle={Proceedings of the 2025 Conference on Empirical Methods in Natural Language Processing},
  pages={15428--15454},
  year={2025}
}

@article{hu2020open,
  title={Open graph benchmark: Datasets for machine learning on graphs},
  author={Hu, Weihua and Fey, Matthias and Zitnik, Marinka and Dong, Yuxiao and Ren, Hongyu and Liu, Bowen and Catasta, Michele and Leskovec, Jure},
  journal={Advances in neural information processing systems},
  volume={33},
  pages={22118--22133},
  year={2020}
}

@inproceedings{wang2019evaluating,
  title={On evaluating embedding models for knowledge base completion},
  author={Wang, Yanjie and Ruffinelli, Daniel and Gemulla, Rainer and Broscheit, Samuel and Meilicke, Christian},
  booktitle={Proceedings of the 4th Workshop on Representation Learning for NLP (RepL4NLP-2019)},
  pages={104--112},
  year={2019}
}

@article{zhang2020improve,
  title={Improve the translational distance models for knowledge graph embedding},
  author={Zhang, Siheng and Sun, Zhengya and Zhang, Wensheng},
  journal={Journal of Intelligent Information Systems},
  volume={55},
  number={3},
  pages={445--467},
  year={2020},
  publisher={Springer}
}

@inproceedings{schlichtkrull2018modeling,
  title={Modeling relational data with graph convolutional networks},
  author={Schlichtkrull, Michael and Kipf, Thomas N and Bloem, Peter and Van Den Berg, Rianne and Titov, Ivan and Welling, Max},
  booktitle={European semantic web conference},
  pages={593--607},
  year={2018},
  organization={Springer}
}

@inproceedings{shang2019end,
  title={End-to-end structure-aware convolutional networks for knowledge base completion},
  author={Shang, Chao and Tang, Yun and Huang, Jing and Bi, Jinbo and He, Xiaodong and Zhou, Bowen},
  booktitle={Proceedings of the AAAI conference on artificial intelligence},
  volume={33},
  number={01},
  pages={3060--3067},
  year={2019}
}

@inproceedings{
Vashishth2020Composition-based,
title={Composition-based Multi-Relational Graph Convolutional Networks},
author={Shikhar Vashishth and Soumya Sanyal and Vikram Nitin and Partha Talukdar},
booktitle={International Conference on Learning Representations},
year={2020},
url={https://openreview.net/forum?id=BylA_C4tPr}
}

@article{dai2022mrgat,
  title={MRGAT: multi-relational graph attention network for knowledge graph completion},
  author={Dai, Guoquan and Wang, Xizhao and Zou, Xiaoying and Liu, Chao and Cen, Si},
  journal={Neural Networks},
  volume={154},
  pages={234--245},
  year={2022},
  publisher={Elsevier}
}

@inproceedings{li2023message,
  title={Are message passing neural networks really helpful for knowledge graph completion?},
  author={Li, Juanhui and Shomer, Harry and Ding, Jiayuan and Wang, Yiqi and Ma, Yao and Shah, Neil and Tang, Jiliang and Yin, Dawei},
  booktitle={Proceedings of the 61st Annual Meeting of the Association for Computational Linguistics (Volume 1: Long Papers)},
  pages={10696--10711},
  year={2023}
}

\appendix

\begin{table}[h]
\centering
\resizebox{\columnwidth}{!}{%
\begin{tabular}{|c|c|}
\hline
Pattern                                                                                                & Constraint Matrices $(E=A_\psi - A_\phi)$                        \\ \hline
\begin{tabular}[c]{@{}c@{}}Hierarchy\\ $r_1(X,Y) \Rightarrow r_2(X,Y)$\end{tabular}                     &               $(A_{r_1}-A_{r_2})X=0$                  \\ \hline

\begin{tabular}[c]{@{}c@{}}Symmetry\\ $r(X,Y) \Rightarrow r(Y,X)$\end{tabular}                         &  $(A_{r}^2-I)X=0$                               \\ \hline
\begin{tabular}[c]{@{}c@{}}Antisymmetry\\ $r(X,Y) \Rightarrow \neg r(X,Y)$\end{tabular}                &   $(A_{r}^2-I)X\neq0$                              \\ \hline
\begin{tabular}[c]{@{}c@{}}Inversion\\ $r_1(X,Y) \Rightarrow r_2(Y, X)$\end{tabular}                   &    $(A_{r_2}A_{r_1}-I)X=0$                             \\ \hline
\begin{tabular}[c]{@{}c@{}}Intersection\\ $r_1(X,Y) \wedge r_2(X,Y) \Rightarrow r_3(X,Y)$\end{tabular} &     $(A_{r_1} - A_{r_3})X = (A_{r_2} - A_{r_3})X = 0 $                            \\ \hline

\begin{tabular}[c]{@{}c@{}}Transitivity\\ $r(X,Y) \wedge r(Y,Z) \Rightarrow r(X,Z)$\end{tabular}       &    $(A_{r}^2 - A_{r})X=0$                             \\ \hline
\begin{tabular}[c]{@{}c@{}}Composition\\ $r_1(X,Y) \wedge r_2(Y,Z) \Rightarrow r_3(X,Z)$\end{tabular}  & $(A_{r_2}A_{r_1} - A_{r_3})X=0$ \\ \hline

\begin{tabular}[c]{@{}c@{}}Gen. Intersection\\ $r_1(X,Y) \wedge r_1(Y,X) \Rightarrow r_2(X,Y)$\end{tabular}  & $(A_{r_1}A_{r_2}-I)X=(A_{r_1}-A_{r_2})X=0 $ \\ \hline

\begin{tabular}[c]{@{}c@{}}B. Transitive\\ $r(Y,Z) \wedge r(Z,X) \Rightarrow r(X,Y)$\end{tabular}  & $(A_{r}^3 -I)X=0 $ \\ \hline

\begin{tabular}[c]{@{}c@{}}Equality\\ $r(X,Z) \wedge r(Y,Z) \Rightarrow r(X,Y)$\end{tabular}  & $ (I -A_{r})X=0$  \\ \hline

\begin{tabular}[c]{@{}c@{}}B. Composition\\ $r_1(Y,Z) \wedge r_2(Z,X) \Rightarrow r_3(X,Y)$\end{tabular}  & $ (A_{r_2}A_{r_1}A_{r_3}-I)X=0$ \\ \hline

\begin{tabular}[c]{@{}c@{}}Commonality\\ $r_1(X,Z) \wedge r_2(Y,Z) \Rightarrow r_3(X,Y)$\end{tabular}  & $ (A_{r_2}^{-1}A_{r_1}-A_{r_3})X=0$ \\ \hline

\end{tabular}}
\caption{Constraint matrices corresponding to various inference patterns. The patterns are presented in~\cite{krishnan2024method}}
\label{tab:constraint matrices}
\end{table}

\begin{table*}[t]
\small
\centering
\begin{tabular}{lccccccc}
\hline
Dataset     & $B$    & $N$     & $D$ & $\gamma$ & $\alpha$ & $lr$ & $k$ \\ \hline
WN18RR      & 512    & 1024  & 1000 & 6.0 & 0.5 & 0.00005 & 20                  \\
FB15k-237     & 1024    & 256  & 1500 & 6.0 & 1.0 & 0.00005 & 20         \\
YAGO3-10 & 1024 & 400 & 1000 & 24.0 & 1.0 & 0.0002 & 2         \\
ogbl-biokg & 512 & 128 & 2000 & 12.0 & 1.0 & 0.001 & 12                \\
ogbl-wikikg2 & 4096 & 250 & 100 & 7.0 & 1.0 & 0.005 & 20                \\ \hline
\end{tabular}
\caption{The best hyperparameter settings of PogRE for link prediction. $B$, $N$, $D$, $\gamma$, $\alpha$, $lr$, and $k$ denote batch size, negative sampling size, dimension, gamma (presented in Equation~\ref{eq:gamma}), alpha (presented in Equation~\ref{eq:alpha}), learning rate, and number of Householder reflections, respectively.}
\label{tab:hyperparams}
\end{table*}

\begin{table*}[t]
\small
\centering
\begin{tabular}{lcccc}
\hline
\multirow{2}{*}{Model}
& \multirow{2}{*}{Relation Parameters}
& \multirow{2}{*}{Required Relation Tensors}
& \multicolumn{2}{c}{Peak GPU Memory During Training} \\
\cline{4-5}
& & & WN18RR & FB15k-237 \\
\hline

RotatE 
& $n_r d$ 
& $b \times d$ 
& 11,384 MB
& 11,391 MB \\

PairRE 
& $2n_r d$ 
& $2(b \times d)$ 
& 14,370 MB
& 10,855 MB \\

PogRE 
& $n_r(k+4)d + \frac{d(d+1)}{2}$ 
& $4(b \times d) + b \times k \times d + \frac{d(d+1)}{2}$ 
& 14,624 MB
& 12,627 MB \\

PogRE w/o QR 
& $n_r(d^2+4d)$ 
& $4(b \times d) + b \times d \times d$ 
& OOM
& OOM \\

\hline
\end{tabular}
\caption{Space complexity comparison of PogRE and baseline models. 
$n_r$, $d$, $k$, and $b$ denote the number of relations, embedding dimension, number of Householder reflections, and batch size, respectively. 
Peak GPU memory consumption is measured during training on WN18RR and FB15k-237 using a single NVIDIA GeForce RTX 3090 under the same experimental settings.}
\label{tab:space_complexity}
\end{table*}

\section{Constraint Matrices for Patterns and Local Pattern Generalization}
\label{appendix:constraint_matrices}

\paragraph{Constraint Matrices for Various Patterns}
Table~\ref{tab:constraint matrices} summarizes the derived constraint matrices for various patterns widely used in KG. Note that for patterns having multiple paths (e.g., Intersection), $E$ represents a set of matrices $\{E_1, E_2, \dots\}$ to be satisfied simultaneously.

\paragraph{Bounding Lipschitz Constants for Local Pattern Generalization to Unobserved Entities}
By applying Spectral Normalization to the shared matrix $R$, we ensure that the spectral norm of each relation-specific linear transformation is bounded: $\|L_r\|_2 \le 1$. Since the constraint matrix $E$ is defined as $L_\psi - L_\phi$, the spectral norm of $E$ is also bounded by the triangle inequality:$$\|E\|_2 = \|L_\psi - L_\phi\|_2 \le \|L_\psi\|_2 + \|L_\phi\|_2 \le 2$$This bound ensures that the transformation defined by the constraint matrix is Lipschitz continuous. For an entity $e_{obs}$ that is known to satisfy the pattern (i.e., $\|Ee_{obs}\| \approx 0$) and a semantically similar but unobserved entity $e_{unobs}$, the pattern error $\|Ee_{unobs}\|$ for $e_{unobs}$ is bounded as follows:
$$\|Ee_{unobs}\| \le \|E\|_2 \|e_{unobs} - e_{obs}\| + \|Ee_{obs}\|$$
As shown in the inequality, if the distance $\|e_{unobs} - e_{obs}\|$ is small, the error $\|Ee_{unobs}\|$ remains small. This mathematically guarantees that the model generalizes the learned pattern from observed entities to semantically similar entities with similar embeddings.

\begin{table*}[t]
\small
\centering
\begin{tabular}{lccccc}
\hline
\multirow{2}{*}{Model}
& \multirow{2}{*}{Time Complexity}
& \multicolumn{2}{c}{Training Time}
& \multicolumn{2}{c}{MRR} \\
\cline{3-6}
& & WN18RR & FB15k-237 & WN18RR & FB15k-237 \\
\hline

RotatE 
& $O(bd)$ 
& 1h 40m 
& 2h 20m 
& .476 
& .338 \\

PairRE 
& $O(bd)$ 
& 2h 20m 
& 3h 
& .413 
& .351 \\

PogRE ($k=2$) 
& $O(bd^2)$ 
& 2h 50m 
& 3h 
& .503 
& .362 \\

PogRE ($k=20$) 
& $O(bd^2)$ 
& 3h 40m 
& 4h 30m 
& .506 
& .369 \\

PogRE (w/o QR) 
& $O(bd^2)$ 
& -- 
& -- 
& -- 
& -- \\

\hline
\end{tabular}
\caption{Time complexity, training time, and link prediction performance of PogRE and baseline models. 
$b$ and $d$ denote the batch size and embedding dimension, respectively. 
Training times are measured using a single NVIDIA GeForce RTX 3090 under the same experimental settings.}
\label{tab:time_complexity}
\end{table*}

\begin{figure*}[t]
    \centering
    
    \subfigure[WN18RR]{
        \includegraphics[width=0.32\textwidth]{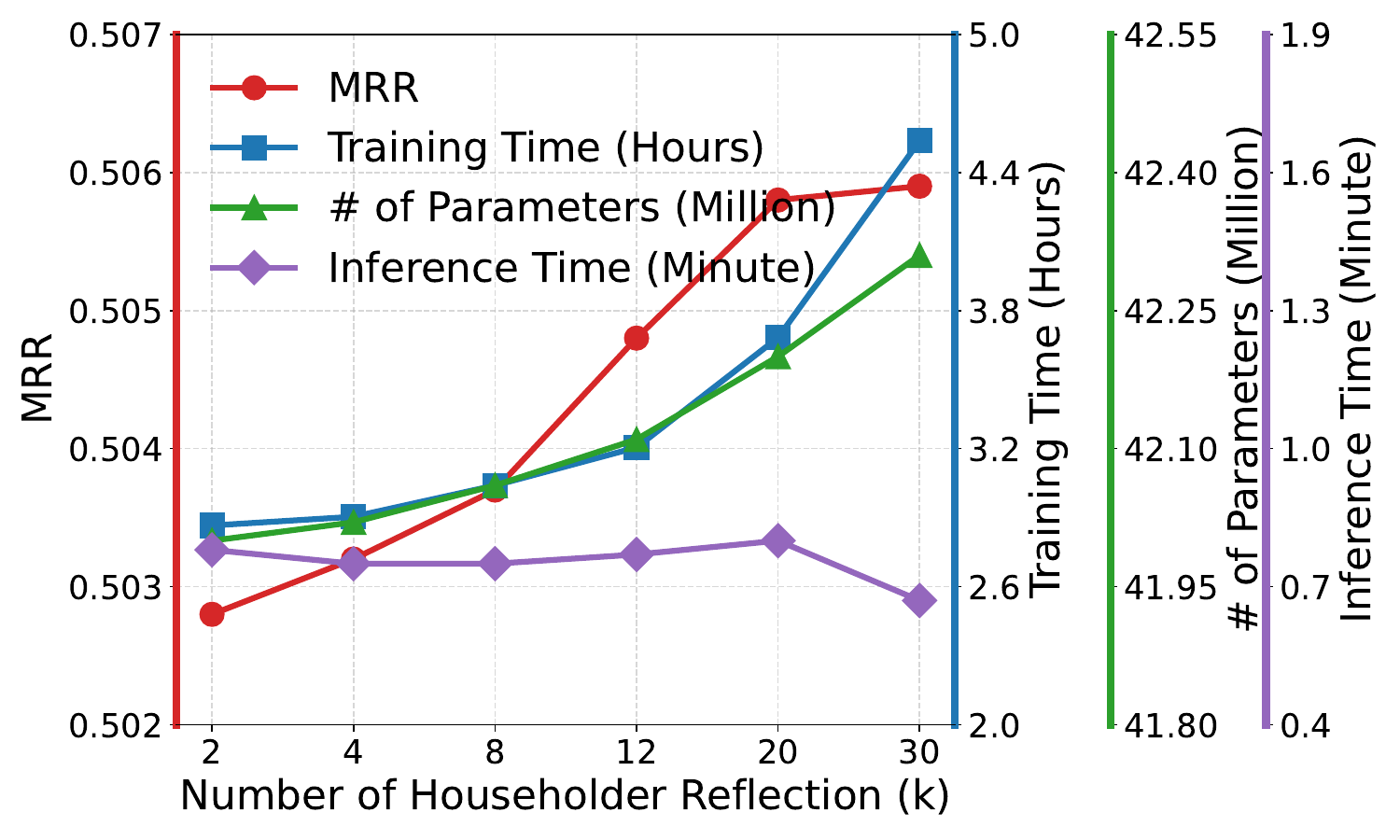}}
        \label{fig:computational_complexity_a}
        \hfill
    \subfigure[FB15k-237]{
        \includegraphics[width=0.32\textwidth]{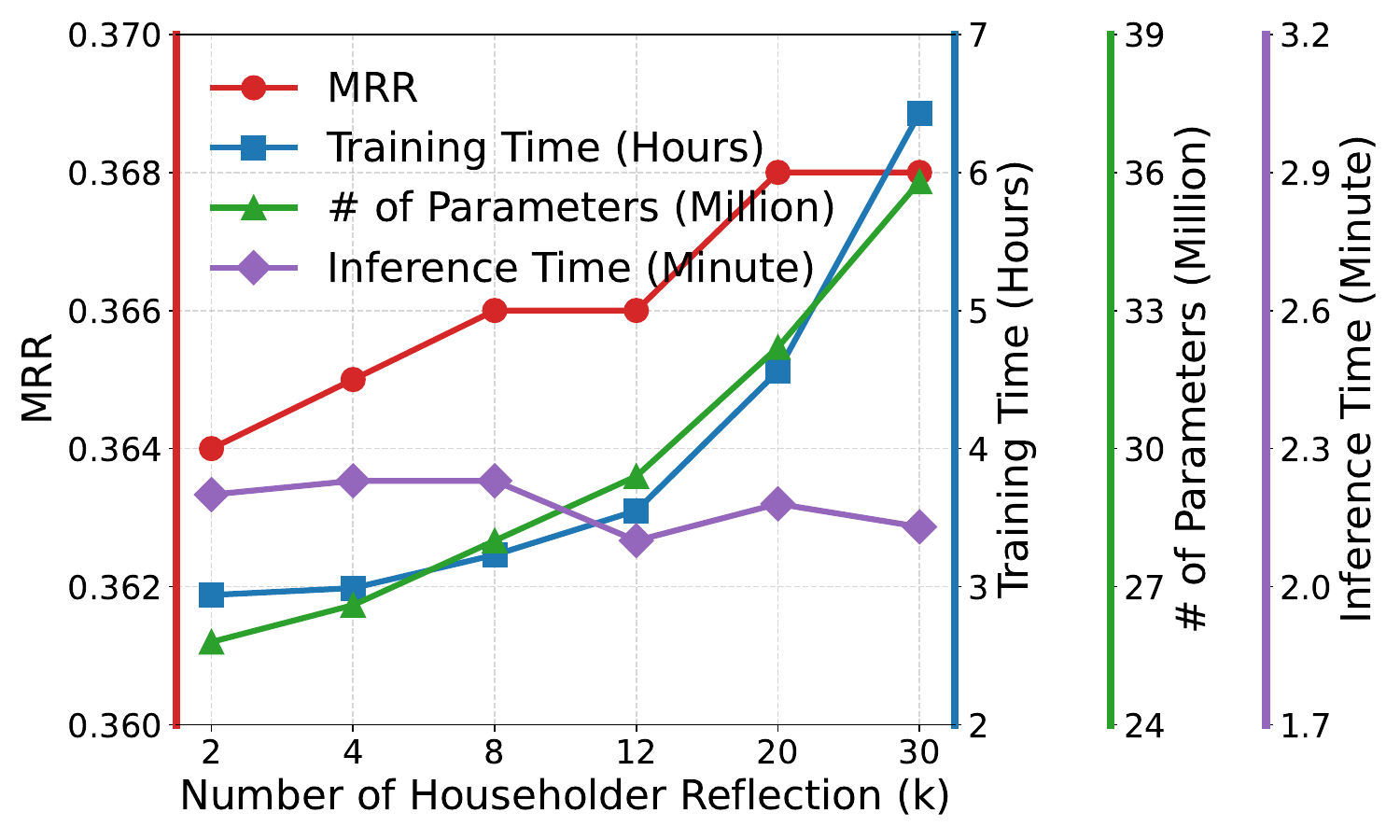}}
        \label{fig:computational_complexity_b}
        \hfill
    \subfigure[YAGO3-10]{
        \includegraphics[width=0.32\textwidth]{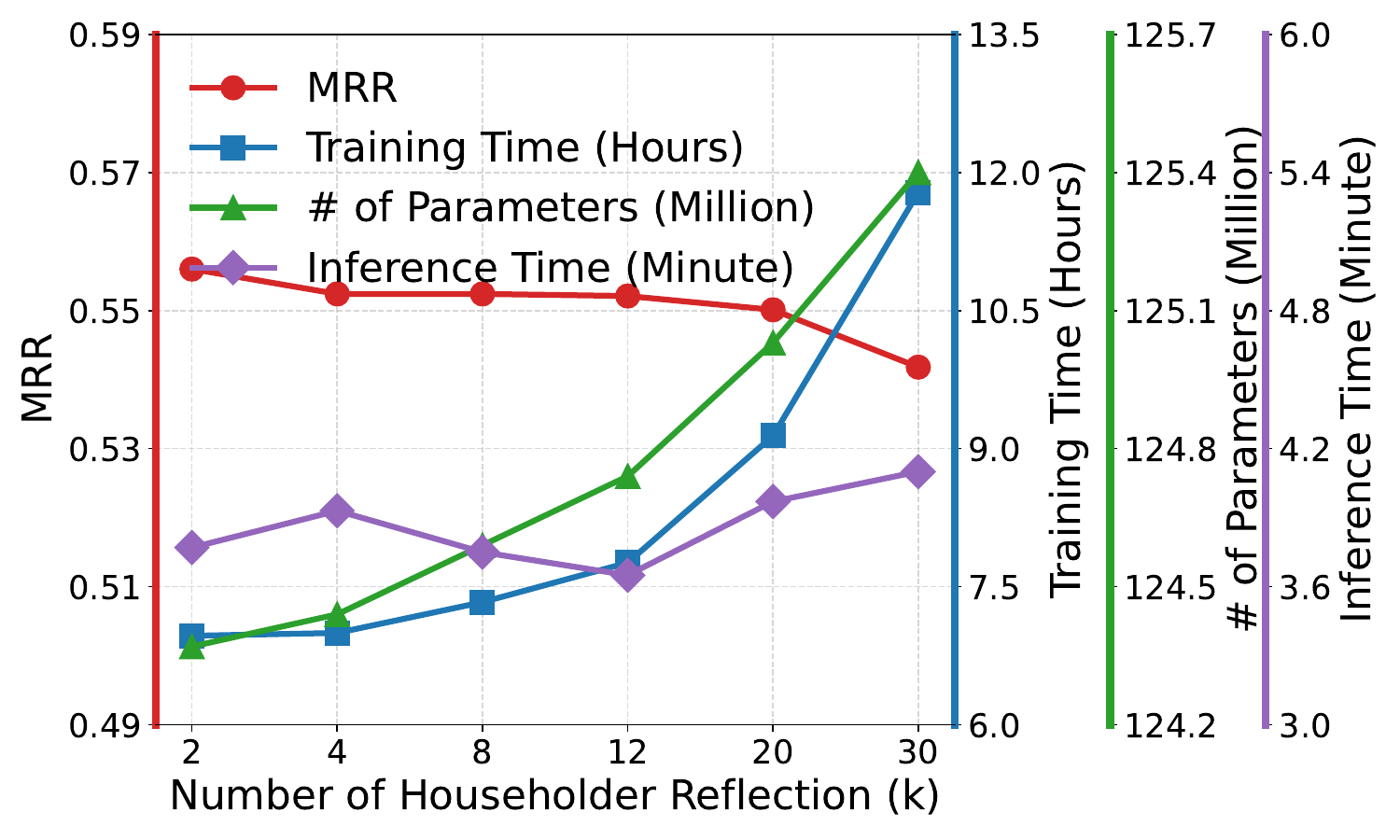}}
        \label{fig:computational_complexity_c}
    \caption{MRR, training time, inference time, and number of parameters of PogRE on three benchmark datasets.}
    \label{fig:computational_complexity}
\end{figure*}

\section{Implementation Details}
For the experiments, we adopt the hyperparameter settings from RotatE~\cite{sun2019rotate} for WN18RR and YAGO3-10, and from PairRE~\cite{chao2021pairre} for FB15k-237. Additionally, for PogRE, the number of Householder reflections $k$ is selected from $\{2, 4, 8, 12, 20\}$. More specifically, we utilized the official implementations of RotatE~\cite{sun2019rotate} and PairRE~\cite{chao2021pairre} as our codebase. For the datasets, we used WN18RR, FB15k-237, and YAGO3-10 as provided in the official RotatE repository, and the biokg and wikikg2 datasets as provided in the PairRE repository. Table~\ref{tab:hyperparams} presents the exact batch size, negative sampling size, embedding dimension, $\gamma$, learning rate, and $k$ used for each dataset. Our presented results represent the mean of three independent runs for each dataset. Furthermore, Scaling $S_r$ and Rotation $R_r$ are used for FB15k-237 and YAGO3-10, whereas Translation $T_r$ and Rotation $R_r$ are used for WN18RR. Finally, following Rotate3D~\cite{gao2020rotate3d}, an $L_2$ regularizer is applied to entity embeddings for WN18RR. Experiments for the PogRE were conducted on an NVIDIA 3090 with 24GB of memory.

\section{Computational Complexity}
\label{sec:computational_complexity}

\paragraph{Space Complexity}
Table~\ref{tab:space_complexity} compares the number of relation parameters, the relation tensors required during batch scoring, and the peak GPU memory consumption during training. $n_r$, $d$, $k$, and $b$ denote the number of relations, embedding dimension, number of Householder reflections, and batch size, respectively.

In PogRE, the relation-specific orthogonal transformation is represented using $k$ Householder vectors, while the upper-triangular matrix is shared across all relations. Therefore, PogRE requires $O(n_rkd+d^2)$ parameters for its dense linear transformations. In contrast, PogRE w/o QR assigns an independent $d \times d$ dense matrix to each relation, resulting in $O(n_rd^2)$ relation parameters. The difference becomes more pronounced during batch scoring. PogRE w/o QR requires a $b \times d \times d$ tensor containing relation-specific dense matrices, whereas PogRE requires $b \times k \times d$ relation-specific Householder vectors and a single shared $d \times (d+1)/2$ matrix. This quadratic memory requirement at the batch level makes PogRE w/o QR infeasible under the same experimental setting and results in OOM.

\paragraph{Time Complexity}
Table~\ref{tab:time_complexity} compares the theoretical scoring complexity and the actual training time. PogRE w/o QR has the same theoretical time complexity as PogRE but is infeasible under the same experimental setting due to its substantially higher space complexity. PogRE has a higher theoretical time complexity than RotatE and PairRE because of the shared matrix multiplication. Nevertheless, its practical training time with $k=2$ remains comparable to that of the baselines, while achieving higher performance in link prediction. Increasing $k$ to 20 requires additional training time but further improves the MRR on both datasets. These results demonstrate that PogRE provides a practical trade-off between computational cost and performance.

Figure~\ref{fig:computational_complexity} presents computational complexity and performance with respect to the Householder reflection $k$. In WN18RR and FB15k-237, performance improves as $k$ increases but shows no significant improvement after $k=20$. This performance gain is accompanied by an increase in computational cost as $k$ grows. In YAGO3-10, the MRR is highest at $k=2$ and decreases as $k$ increases. These results suggest that while a larger $k$ can improve performance by increasing the expressive power, excessive complexity may lead to a decrease in performance due to overfitting. Furthermore, there is almost no variation in inference time across different $k$ values, implying that $k$ can be selected during training without concerns regarding inference time.

\section{Impact of Over-generalization}

\label{appendix:over-generalization_experiments}

\paragraph{Detailed Definition of Group $G_{over}$} Group $G_{over}$ consists of test triples $(h, r, t)$ where at least one candidate triple $(h, r, t')$ (where $t' \neq t$) is $\phi_i \in G_u \setminus G_o$ for which there is a corresponding $\psi_i \in G_o$. In this case, the body instances corresponding to the candidate appear in the training set. For instance, consider a local composition pattern $\psi \Rightarrow\phi$ consisting of the relation triplet $(r_1, r_2, r)$. If the training set contains the body instances $(h, r_1, x)$ and $(x, r_2, t')$ for at least one candidate $t'$ and some entity $x \in E$, then the test triple $(h, r, t)$ is assigned to $G_{over}$. If KGE models suffer from over-generalization, they are likely to assign a high score to such a candidate $(h, r, t')$, treating it as a valid triple. For simplicity, we only describe the case of tail prediction, but the same procedure applies to head prediction. 
For pattern, we consider symmetry, inversion, composition, hierarchy, intersection, transitive, g.intersection, b. transitive and b. composition patterns, as they are the most representative inference patterns extensively investigated across a wide range of KGE models. Antisymmetry is not considered because it ensures the absence of a head, rather than ensuring the presence of head.

\begin{figure}[t]
    \centering
    
    \subfigure[WN18RR]{
        \includegraphics[width=1\columnwidth]{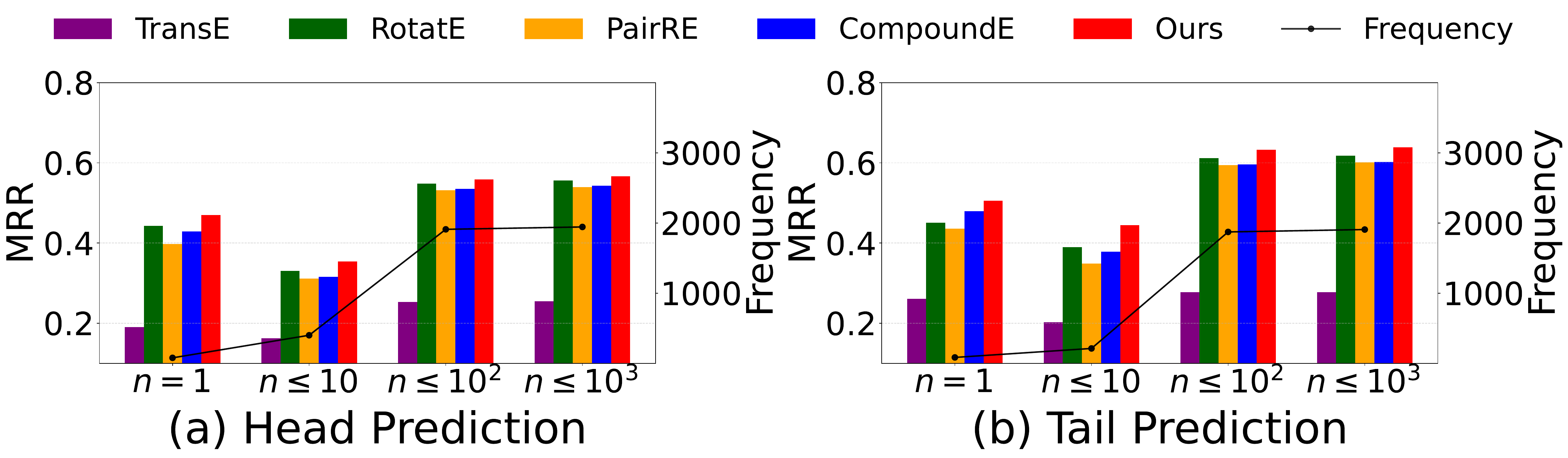}
    }
    
    \vspace{1em} 
    
    \subfigure[YAGO3-10]{
        \includegraphics[width=1\columnwidth]{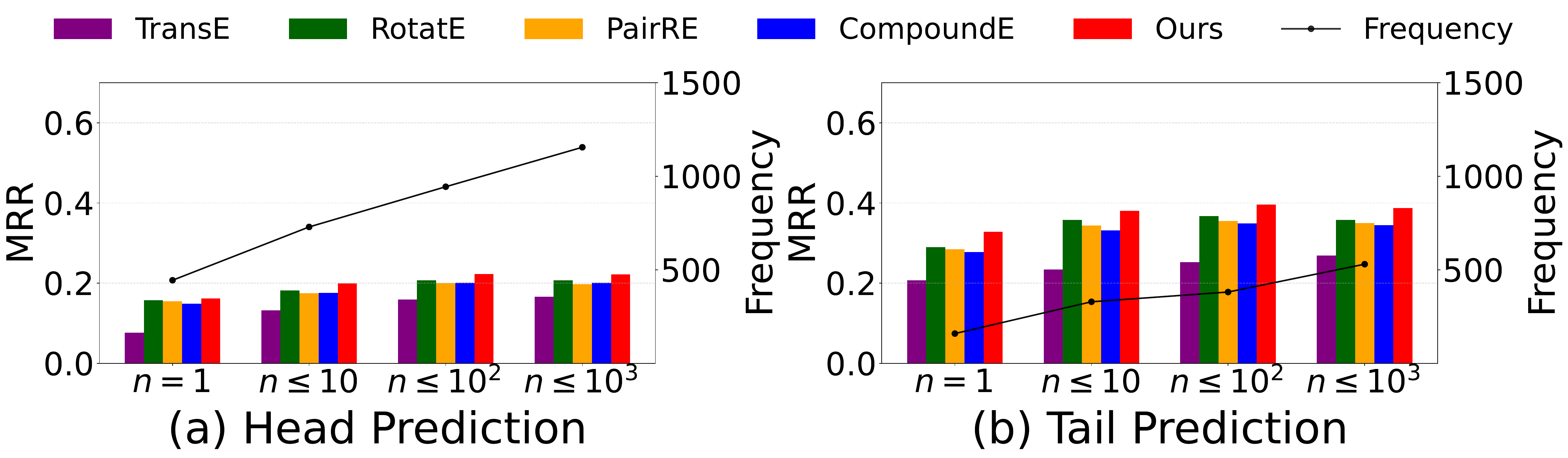}
    }
    
    \vspace{1em} 
    
    \caption{MRR comparison between PogRE and baseline models for various sparsity conditions of pattern instances on two benchmarks: WN18RR (a), and YAGO3-10 (b). The black line indicates the number of test triples for each group.}
    \label{fig:composition_experiments_all}
\end{figure}

\begin{table*}[t]
\small
\centering
\begin{tabular}{lllllll}
\hline
\multirow{2}{*}{Model} & \multicolumn{3}{c}{ogbl-biokg}       & \multicolumn{3}{c}{ogbl-wikikg2}    \\ \cline{2-7} 
\multicolumn{1}{c}{}                       & \multicolumn{1}{c}{Dim}  & \multicolumn{1}{c}{Valid MRR}     & \multicolumn{1}{c}{Test MRR}      & \multicolumn{1}{c}{Dim} & \multicolumn{1}{c}{Valid MRR}     & \multicolumn{1}{c}{Test MRR}      \\ \hline
TransE                                     & 2000 & 0.7456$\pm$0.0003 & 0.7452$\pm$0.0004 & 500 & 0.4272$\pm$0.0030 & 0.4256$\pm$0.0030 \\
DistMult                                   & 2000 & 0.8055$\pm$0.0003 & 0.8043$\pm$0.0003 & 500 & 0.3506$\pm$0.0042 & 0.3729$\pm$0.0045 \\
ComplEx                                    & 1000 & 0.8105$\pm$0.0001 & 0.8095$\pm$0.0007 & 250 & 0.3759$\pm$0.0016 & 0.4027$\pm$0.0027 \\
RotatE                                     & 1000 & 0.7997$\pm$0.0002 & 0.7989$\pm$0.0004 & 250 & 0.4353$\pm$0.0028 & 0.4332$\pm$0.0025 \\
PairRE                                     & 2000 & \underline{0.8172}$\pm$0.0005 & \underline{0.8164}$\pm$0.0005 & 200 & 0.5423$\pm$0.0020 & 0.5208$\pm$0.0027 \\
ComopoundE                                 & \multicolumn{1}{c}{-}    & \multicolumn{1}{c}{-}             & \multicolumn{1}{c}{-}             & 100 & \underline{0.6716}$\pm$0.0009        & \underline{0.6473}$\pm$0.0022        \\ \hline
PogRE                                       & 2000 & \textbf{0.8198}$\pm$0.0001 & \textbf{0.8190}$\pm$0.0003 & 100 & \textbf{0.6841}$\pm$0.0010 & \textbf{0.6606}$\pm$0.0006 \\ \hline
\end{tabular}

\caption{Link prediction results on large-scale KGs, including ogbl-biokg and ogbl-wikikg2. Bold indicates the best result, and underline indicates the second best result. $\pm$ indicates the standard deviation.}
\label{tab:large-scale_MRR}
\end{table*}

\paragraph{Quantified Impact of Over-generalization}
Figure~\ref{fig:composition_experiments_all} presents the performance comparison on the $G_{over}$ across WN18RR and YAGO3-10. Also in WN18RR and YAGO3-10, PogRE consistently outperforms the baselines regardless of the local pattern criteria and the dataset. These results demonstrate that PogRE effectively addresses the negative impact of over-generalization, and shows robustness across datasets.

\paragraph{Distribution of Patterns by Number of Pattern Instances}
Table~\ref{tab:integrated_patterns} presents the distribution of each pattern according to the number of its pattern instances ($n$) across three KG benchmark datasets.In WN18RR, FB15k-237, and YAGO3-10, respectively, 20.0\%, 22.7\%, and 41.7\% of symmetry patterns, 27.3\%, 24.6\%, and 20.0\% of antisymmetry patterns, and 50.0\%, 49.3\%, and 58.8\% of inversion patterns have 10 or fewer pattern instances. This result indicates that for symmetry, antisymmetry, and inversion, a substantial proportion of patterns in real-world KGs are observed in only a few instances.

\begin{table}[t]
    \centering
    \tiny
    \begin{tabular}{cccccc}
    \hline
    \multirow{2}{*}{\textbf{Dataset}} &
    \multirow{2}{*}{\textbf{Entities}} &
    \multirow{2}{*}{\textbf{Relations}} &
    \multicolumn{3}{c}{\textbf{Triples}}\\
        \cline{4-6}
     & & & \textbf{Train} & \textbf{Valid} & \textbf{Test} \\
    \hline
    WN18RR & 40,943 & 11 & 86,835 & 3,034 & 3,134 \\
    FB15k-237 & 14,541 & 237 & 272,115 & 17,535 & 20,466 \\
    YAGO3-10 & 123,182 & 37 & 1,079,040 & 5,000 & 5,000\\
    ogbl-biokg & 93,773 & 51 & 4,763,814 & 162,870 & 162,886 \\
    ogbl-wikikg2 & 2,500,604 & 535 & 16,109,182 & 429,456 & 598,543 \\
    \hline
    \end{tabular}
    \caption{Statistics of three benchmark datasets}
    \label{tab:dataset}
\end{table}

\section{Datasets}
\label{appendix:datasets}
WN18RR, FB15k-237 and YAGO3-10 are used to evaluate PogRE. WN18RR and FB15k-237 are subsets of WN18~\cite{bordes2013translating} and FB15k~\cite{bordes2013translating} with inverse relations removed, and YAGO3-10 is a subset of YAGO3~\cite{mahdisoltani2013yago3} containing only entities with a minimum of 10 relations each. ogbl-biokg~\cite{hu2020open} is a large-scale biomedical KG, and ogbl-wikikg2~\cite{hu2020open} is a Wikidata knowledge graph that contains a large number of triples. The statistics are summarized in Table~\ref{tab:dataset}.

\section{Link Prediction Performance on Large-scale Knowledge Graph}
\label{appendix:large-scale_KG}
To verify the effectiveness of PogRE in large-scale KGs, we conduct additional experiments on biokg and wikikg2. As shown in the table~\ref{tab:dataset}, biokg and wikikg2 contain 4.7 million and 16.1 million triples, respectively, making them significantly larger than the standard benchmarks. The comparative results on biokg and wikikg2 are presented in Table~\ref{tab:large-scale_MRR}. PogRE achieves the highest performance in both valid and test MRR compared to the baselines. For instance, PogRE achieves MRR improvements of 0.0027 and 0.0133 over the second-best models, PairRE and CompoundE, on biokg and wikikg2 in Test MRR, respectively. Furthermore, PogRE exhibits a low standard deviation ($\pm 0.001$) across all datasets and settings, indicating that it consistently maintains stable performance regardless of initialization. This comparison demonstrates the robustness of our method across different datasets, especially in large-scale KGs.

\begin{table}[t]
\scriptsize
\centering
\begin{tabular}{lccc}
\hline
\multirow{2}{*}{\begin{tabular}[l]{@{}l@{}}Number of \\ Sample Vector\end{tabular}} & \multicolumn{3}{c}{Rank}                                                                   \\ \cline{2-4} 
                       & \multicolumn{1}{l}{WN18RR} & \multicolumn{1}{l}{FB15k-237} & \multicolumn{1}{l}{YAGO3-10} \\ \hline 
100    & 100.0       & 100.0                      & 100.0                                                 \\
200    & 200.0       & 200.0                      & 200.0                                                 \\
500    & 500.0       & 500.0                      & 500.0                                                \\
1,000  & 996.1$\pm$2.0      &  1000.0                   & 999.4$\pm$0.5                                               \\
1,500  & 1000.0       &   1492.8$\pm$1.2                 & 1000.0                                               \\
2,000  & 1000.0       & 1500.0                     & 1000.0                                               \\ \hline 
\end{tabular}

\caption{Mean rank of the subspace spanned by randomly sampled entities in the specific pattern, over 100 random trials across three benchmarks. The entity dimensions of PogRE are 1,000 on WN18RR and YAGO3-10, and 1,500 on FB15k-237.}
\label{tab:space_span_specific_patterns}
\end{table}

\section{Linear Independence of Entities Observed Within a Pattern}
\label{appendix:linear_independence_within_pattern}
In Section~\ref{sec:experiment_independent}, we empirically observed that the dimension of the space spanned by entities randomly sampled from the entire KG is approximately equal to the number of sampled entities. In this section, we further investigate whether entities observed within specific patterns—rather than across the entire KG—are also linearly independent of each other. Our experimental procedure is as follows: First, we randomly select a pattern containing at least 2,000 instances. We then randomly sample $N$ entities from this pattern (varying $N$ from 100 to 2,000) and measure the rank of the space spanned by these sampled entities. Finally, we repeat this process for 100 independent trials and report the average rank in Table~\ref{tab:space_span_specific_patterns}. Our results demonstrate that the rank of the space spanned by these entities is approximately equal to the number of sampled entities, indicating that linear independence is indeed preserved even within specific patterns.

\section{Detailed Definition of the OG Ratio and Its Comparison on Various Patterns}
\label{appendix:OG_Ratio}
In this section, we provide a detailed definition of the OG Ratio and additional experiments on the OG Ratio.

\begin{table}[t]
\centering
\scriptsize
\begin{tabular}{llll}
\hline
\multicolumn{1}{l}{\multirow{2}{*}{Model}} & \multicolumn{3}{c}{Over-Generalization Ratio} \\ \cline{2-4} 
\multicolumn{1}{c}{}                        & $\mathbf{n \leq 10}$      & $\mathbf{n \leq 10^2}$          & $\mathbf{n \leq 10^3}$      \\ \hline
TransE          	                           & .851       & .850     &   .795     \\
RotatE                                     &   .866     & .852     &   .793     \\
PairRE                                    &   .749      & .719      &   .651     \\
CompoundE                                 &  .744       & .711      &   .638     \\ \hline
PogRE                                       &    \textbf{.707}    & \textbf{.681}      &  \textbf{.618}      \\ \hline
\end{tabular}
\caption{OG ratio comparison between KGE baselines and PogRE on FB15k-237. $n$ indicates the number of pattern instances.}
\label{tab:OG_ratio}
\end{table}

\paragraph{Detailed Definition of OG Ratio}
Since KGE is a relative distance-based method, if the score of a triple $(h,r,t)$ is closer to 0 than that of another triple $(h, r, t')$, it implies that the model considers $(h,r,t)$ to be more plausible than $(h, r, t')$. Based on this relative property, we define the OG ratio as follows. First, we extract head instances corresponding to the body instances of patterns in the KG. We then divide them into two groups—True triples and False triples—and measure the average score of each group (we exclude head instances that are in the training data, as they are used to train the models). Next, we define the OG ratio as the average score of True triples divided by that of False triples. An OG ratio $\approx$ 1 indicates that the model suffers from over-generalization, as it assigns similar scores to both True and False triples. An OG ratio $\approx$ 0 implies that the model effectively avoids over-generalization by assigning higher scores to False triples compared to True triples.

\paragraph{OG Ratio Comparison on Various Patterns}
We use the OG ratio to verify whether our model effectively addresses over-generalization in the entire KG. Specifically, we extract True and False triples for patterns whose number of pattern instances $n$ satisfies $n \le 10$, $n \le 10^2$, and $n \le 10^3$ in FB15k-237, and compare the OG ratio of PogRE against KGE baselines, including TransE, RotatE, PairRE, and CompoundE. Table~\ref{tab:OG_ratio} presents a comparison of the OG ratios between PogRE and the KGE baselines. Our model exhibits a lower OG ratio than all other models across all settings. This demonstrates that our model effectively addresses pattern over-generalization in the entire KG. This analysis was not conducted for WN18RR and YAGO3-10; due to their small validation/test sets and data sparsity, these datasets contain very few True triples (fewer than 10).

\section{Empirical Analysis of Local and Universal Patterns}
\label{appendix:empirical_local_global}
In Table~\ref{tab:relation_symmetry}, we present examples of local and universal patterns in real-world KGs based on human verification. Because symmetry and inversion patterns can be easily classified as either universal or local, we extract these patterns from three benchmark datasets: WN18RR, FB15k-237, and YAGO3-10. For FB15k-237, due to the large number of patterns, we randomly extract only 40 patterns. We then manually classify each pattern into two groups: semantically universal and semantically local. We further categorize these patterns based on the number of pattern instances ($n$). Empirically, we observe that, in general, patterns that have low frequencies ($n \le 10$) tend to be semantically local, which can cause over-generalization in existing KGE models, leading to erroneous predictions.

\begin{table}[t]
\centering
\tiny
\begin{tabular}{ll}
\hline
\textbf{Universal but Low Frequency} & \textbf{Local but High Frequency} \\
\hline
\_similar\_to (74, WN) & isLocatedIn (5742, YAGO) \\
\_also\_see (828, WN) & isLocatedIn, hasCapital (1743, YAGO) \\
dealsWith (160, YAGO) & \\
hasNeighbor (550, YAGO) & \\
.../legislative\_sessions (668, FB) & \\
.../military\_combatant\_group (620, FB) & \\
.../canoodled/participant (368, FB) & \\
.../marriage/spouse (342, FB) & \\
.../friendship/friend (204, FB) & \\
.../friendship/participant (1216, FB) & \\
.../dated/participant (1134, FB) & \\
\hline
\end{tabular}
\caption{Representative examples of patterns categorized into the two failure case sets: universal but low frequency, and local but high frequency. The numbers in parentheses indicate the frequency of pattern instances, while WN, FB, and YAGO denote the WN18RR, FB15k-237, and YAGO3-10, respectively.}
\label{tab:failure_case_examples}
\end{table}

However, we also note that, due to the semantic complexity of real-world KGs, this general tendency may not always hold. We identify two representative failure cases for our proposed method.
\begin{itemize}
    \item First, \textbf{universal but low-frequency patterns} may appear when semantically universal patterns are observed in only a few instances due to dataset sparsity. In this case, PogRE may fail to generalize them universally.
    \item Second, \textbf{local but high-frequency patterns} may appear when semantically local patterns have many observed instances. In this case, PogRE may generalize them universally, resulting in inappropriate generalization.
\end{itemize}

Table~\ref{tab:failure_case_examples}, a subset of Table~\ref{tab:relation_symmetry}, shows examples of patterns in these two case sets. Each pattern is reported with its frequency and dataset. The universal but low-frequency set indicates semantically universal patterns whose frequency is lower than the entity embedding dimension. The local but high-frequency set indicates semantically local patterns whose frequency is higher than the entity embedding dimension.

We compare PogRE with KGE baselines on these cases. Specifically, for each dataset, we extract test triples that contain relations included in each failure case set and measure MRR. Table~\ref{tab:universal_low_freq} reports the MRR results for the universal but low frequency set. PogRE underperforms compared to CompoundE on FB15k-237 and YAGO3-10. This demonstrates the negative impact of the universal but low frequency failure case on performance of PogRE.

\begin{table}[t]
\centering
\small
\begin{tabular}{lccc}
\hline
\textbf{Models} & \textbf{WN18RR} & \textbf{FB15k-237} & \textbf{YAGO3-10} \\
\hline
TransE    & 0.242 & 0.103 & 0.213 \\
RotatE    & 0.661 & 0.121 & \textbf{0.377} \\
PairRE    & 0.657 & 0.125 & 0.303 \\
CompoundE & 0.660 & \textbf{0.166} & 0.301 \\
\hline
PogRE      & \textbf{0.686} & 0.150 & 0.283 \\
\hline
\end{tabular}
\caption{MRR comparison between PogRE and baseline KGE models for the universal but low frequency case set across WN18RR, FB15k-237, and YAGO3-10. Bold indicates the best performance.}
\label{tab:universal_low_freq}
\end{table}

\begin{table}[t]
\centering
\small
\begin{tabular}{lc}
\hline
\textbf{Models} & \textbf{MRR} \\
\hline
TransE    & 0.088 \\
RotatE    & 0.131 \\
PairRE    & 0.155 \\
CompoundE & \textbf{0.185} \\
\hline
PogRE      & 0.172 \\
\hline
\end{tabular}
\caption{MRR comparison between PogRE and baseline KGE models for the local but high frequency case set. Note that the evaluated patterns for this case are observed exclusively within YAGO3-10. Bold indicates the best performance.}
\label{tab:local_high_freq}
\end{table}

\begin{table*}[t]
\small
\centering
\begin{tabular}{lccc}
\hline
Pattern Type      & \# of PI    & \# of BI     & Ratio (PI/BI) \\ \hline
Local Pattern \textit{(film/written\_by, actor/film, film/prequel)}     & 13    & 74     & 0.176         \\
Local Pattern \textit{(film/director, film/prequel, actor/film)}     & 22    & 1,895  & 0.012         \\
Universal Pattern \textit{(actor/film, film/country, people/nationality)} & 5,033 & 11,659 & 0.432         \\
Universal Pattern \textit{(people/place\_of\_birth, location/country, people/nationality)} & 1,082 & 1,422  & 0.761         \\ \hline
\end{tabular}
\caption{Comparison of Pattern Instances (PI) and Body Instances (BI) between the local and universal patterns introduced in Figure~\ref{fig:problem}. '\# of PI' and '\# of BI' represent the number of pattern instances and body instances, respectively.}
\label{tab:PI_BI_comparison}
\end{table*}

Interestingly, PogRE still outperforms CompoundE on WN18RR. We analyze the reasons for this as follows. For WN18RR, we measured the MRR for two patterns: \textit{\_similar\_to} and \textit{\_also\_see}. First, despite the \textit{\_similar\_to} pattern having only 74 instances, all models except TransE achieved an MRR of 1.0. This pattern is a potential failure case, but not empirically harmful because of dataset-specific or relatively easy test structure. Furthermore, the \textit{\_also\_see} pattern has 828 instances. Although these instances may not span the entire embedding space, they still provide observed instances that can support generalization. As theoretically shown in Appendix~\ref{appendix:constraint_matrices}, our spectral normalization can help keep the pattern error bounded for unobserved entities if they are semantically similar to the observed entities. Therefore, PogRE can achieve generalization to semantically similar entities for \textit{\_also\_see}, which can explain its superior performance on WN18RR.

Table~\ref{tab:local_high_freq} reports the MRR results for the local but high-frequency set (only evaluated on YAGO3-10). In this case, PogRE also does not outperform CompoundE. This result indicates that PogRE may underperform existing models when semantically local patterns have high frequencies.

Through this analysis, we clarify the applicability boundary of PogRE.
\begin{itemize}
    \item Our method is designed to prevent the over-generalization of local patterns.
    \item As shown in Table~\ref{tab:relation_symmetry}, most local patterns have low frequency, and therefore PogRE can generally improve performance on the overall datasets.
    \item However, PogRE may underperform existing models in failure cases where pattern frequency does not match pattern semantics, such as universal but low-frequency patterns or local but high-frequency patterns.
    \item Therefore, PogRE is most suitable for KGs where low-frequency patterns are likely to be local and high-frequency patterns are likely to be universal.
\end{itemize}

\begin{figure}[t]
    \centering
    
    \includegraphics[width=0.48\columnwidth]{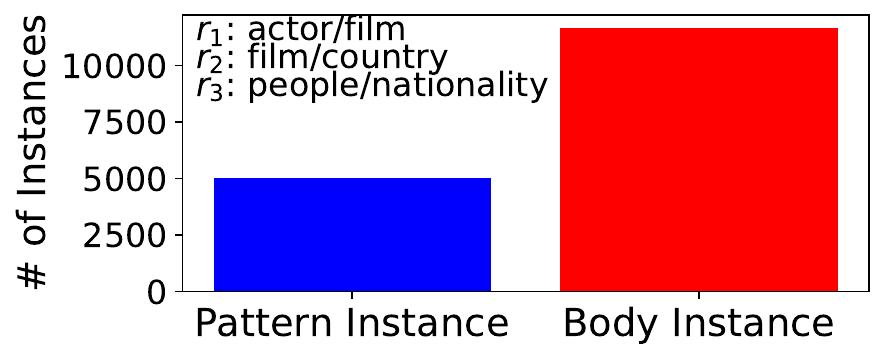}
    \label{fig:PI_BI_c}
    \hfill
    \includegraphics[width=0.48\columnwidth]{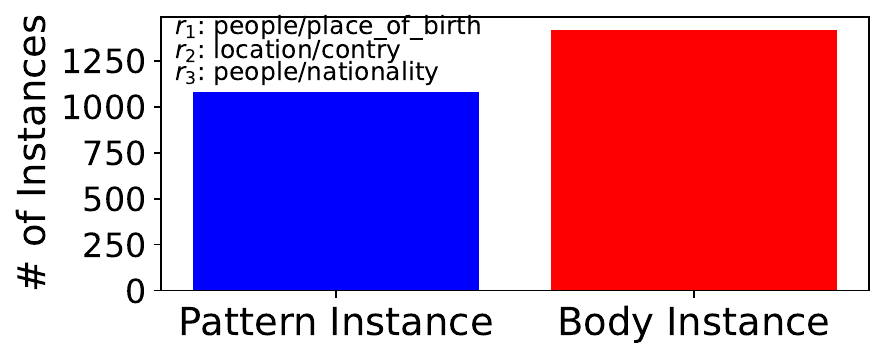}
    \label{fig:PI_BI_d}

    \caption{The number of pattern instances and body instances for the universal patterns introduced in Figure~\ref{fig:problem b}.}
    \label{fig:PI_BI_universal}
\end{figure}

\section{Additional Empirical Evidence for Over-Generalization}
\label{appendix:PI_BI_comparsion}

\paragraph{Comparison of the Number of Pattern and Body Instances Between Local and Universal Patterns} In Figure~\ref{fig:PI_BI_universal}, we present the number of pattern and body instances for the universal patterns introduced in Figure~\ref{fig:problem b}. Additionally, Table~\ref{tab:PI_BI_comparison} compares the PI and BI of the local and universal patterns from Figure~\ref{fig:problem}. As shown in Figure~\ref{fig:PI_BI_universal}, the universal patterns are supported by 5,033 and 1,082 pattern instances, respectively. This significantly exceeds the number of pattern instances of local patterns (13 and 22), demonstrating that universal patterns are supported by a much larger number of observations. Furthermore, as shown in Table~\ref{tab:PI_BI_comparison}, universal patterns exhibit higher PI/BI ratios (0.432 and 0.761) compared to local patterns (0.176 and 0.012). This demonstrates that the patterns in Figure~\ref{fig:problem b} exhibit universal characteristics.

\begin{figure}[t]
    \centering
    
    \subfigure[Histograms of local patterns that are supported by scarce pattern instances. The relations for the left and right figures are \textit{(.../gardening\_hint/split\_to)} and \textit{(.../us\_county/county\_seat)}, respectively.]{
        \includegraphics[width=0.46\columnwidth]{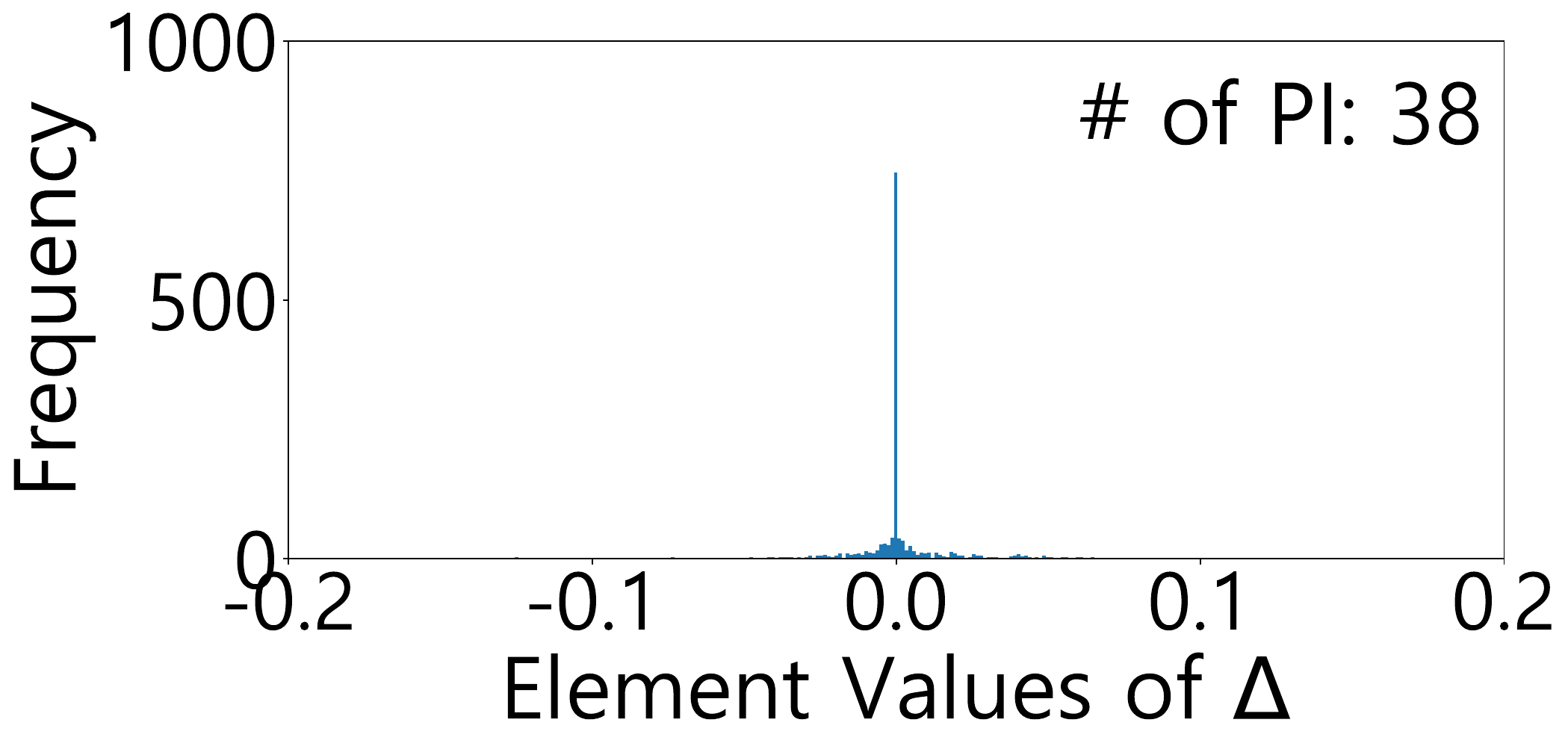}
        \hfill
        \includegraphics[width=0.46\columnwidth]{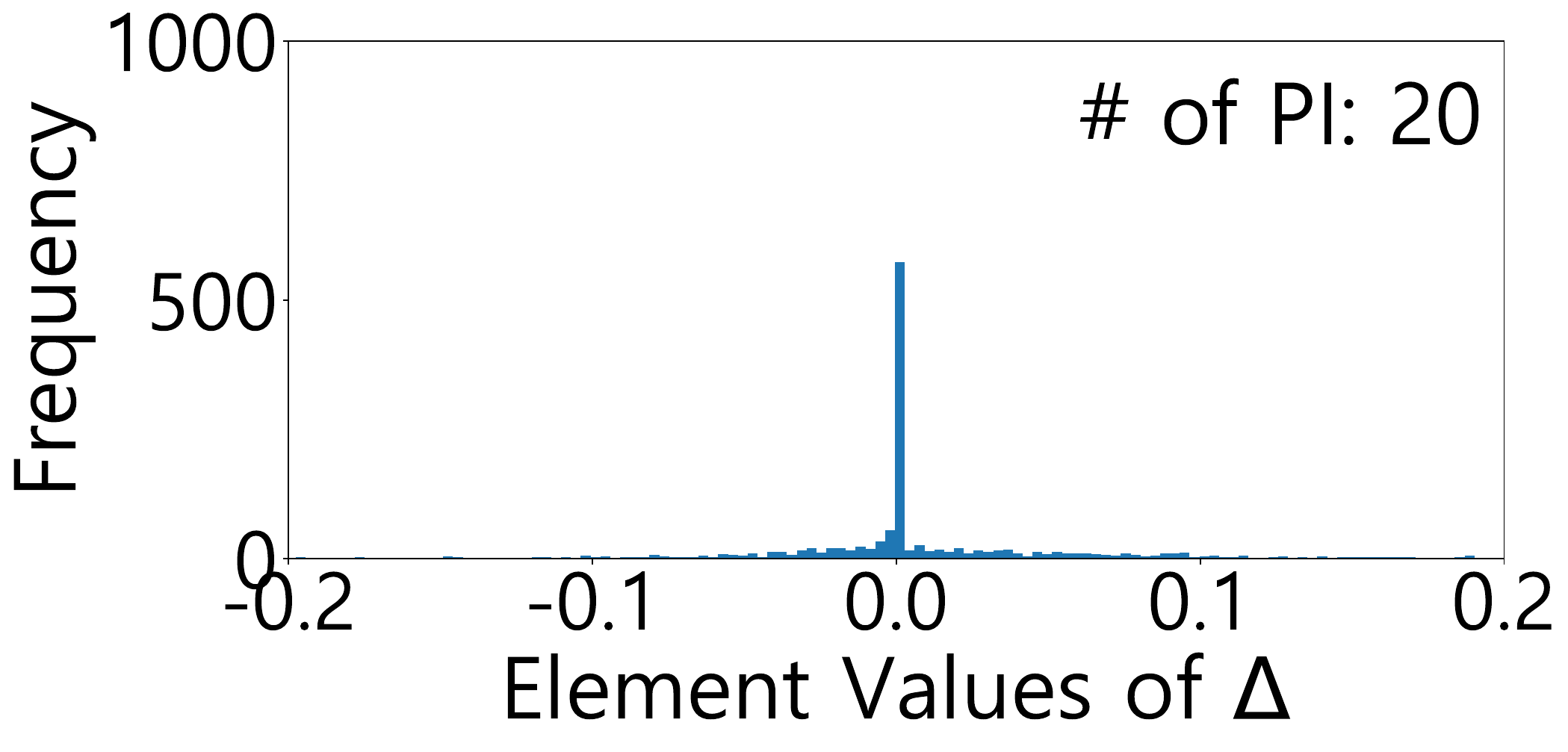}
        \label{fig:problem a symmetry}
    }
    
    \vspace{1em} 
    
    \subfigure[Histograms of universal patterns that are supported by many pattern instances. The relations for the left and right figures are \textit{(.../location/adjoining\_relationship...)} and \textit{(.../award/award\_nomination...)}, respectively.]{
        \includegraphics[width=0.46\columnwidth]{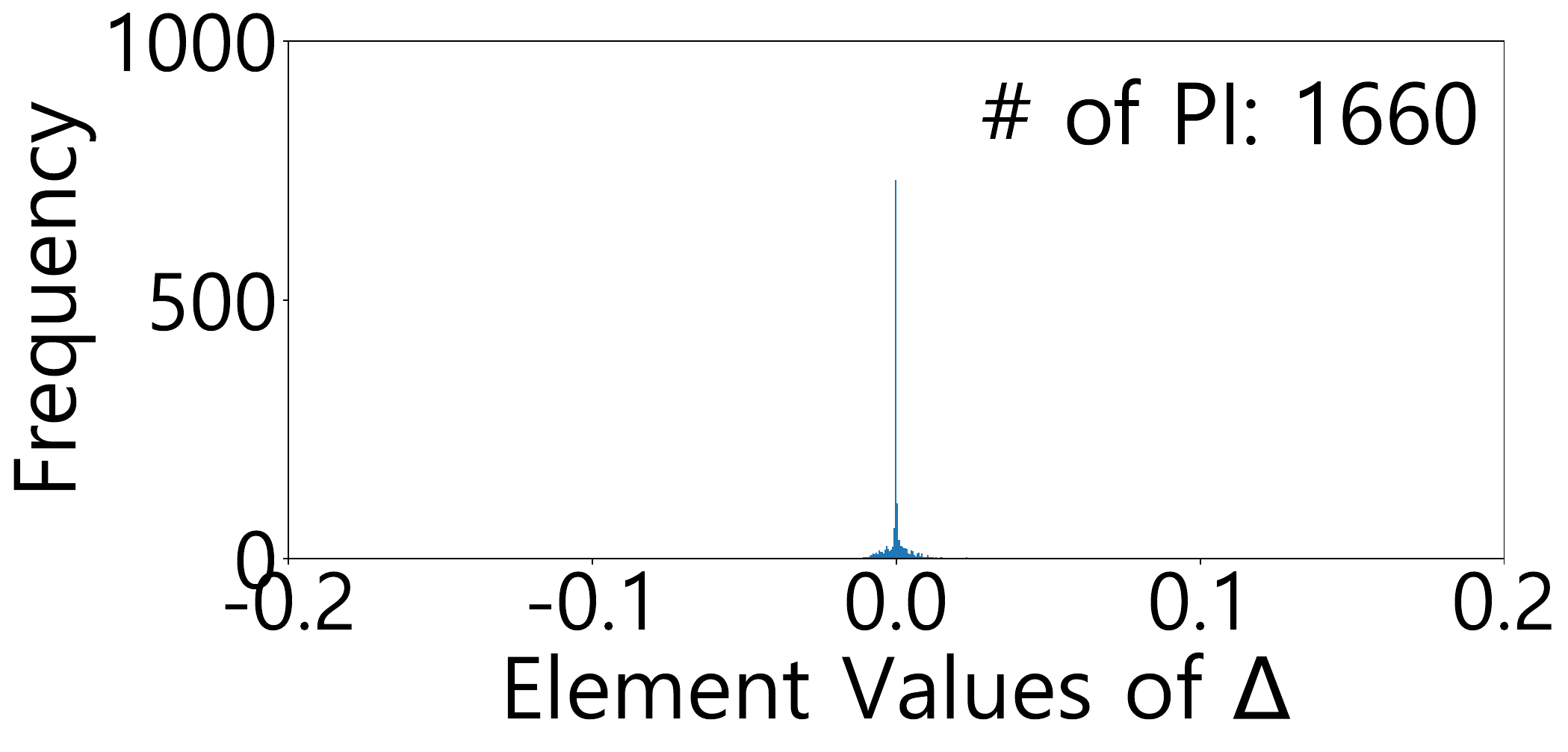}
        \hfill
        \includegraphics[width=0.46\columnwidth]{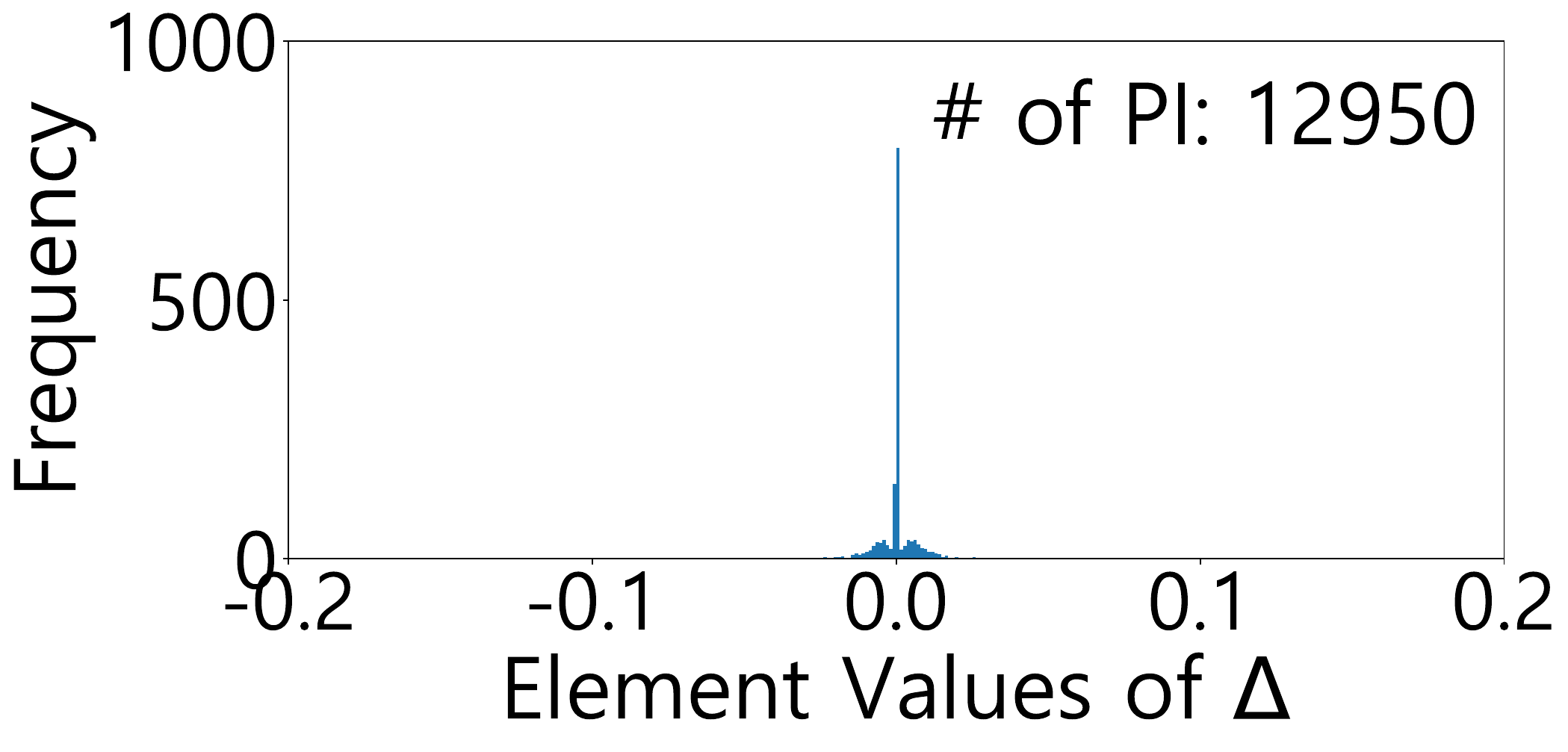}
        \label{fig:problem b symmetry}
    }
    
    \caption{Histograms of embedding difference $\Delta = (r_1^H)^2 - (r_1^T)^2$ for different symmetric relations $r_1$. \# of PI denotes the number of pattern instances. $r_1$ are retrieved from FB15k-237.}
    \label{fig:problem_symmetry}
\end{figure}

\begin{figure}[t]
    \centering
    
    \subfigure[The number of pattern instances and body instances for the local patterns introduced in Figure~\ref{fig:problem a symmetry}.]{
        \includegraphics[width=0.5\columnwidth]{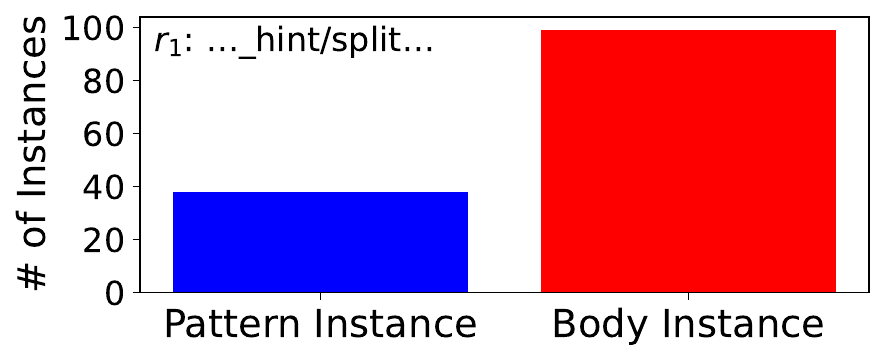}
        \hfill
        \includegraphics[width=0.5\columnwidth]{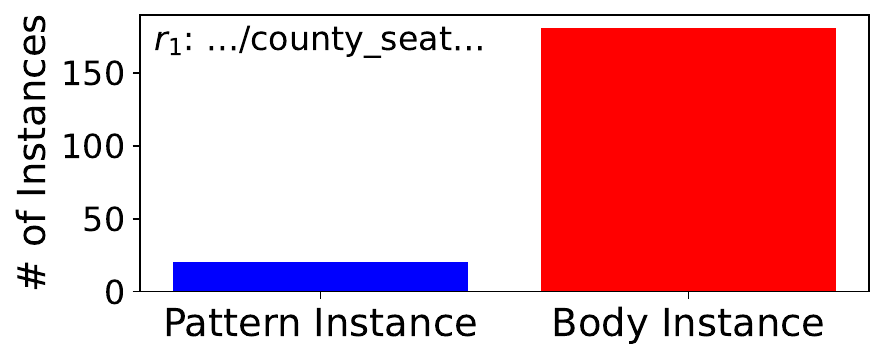}
        \label{fig:PI_BI_local_symmetry}
    }
    
    \vspace{1em} 
    
    \subfigure[The number of pattern instances and body instances for the universal patterns introduced in Figure~\ref{fig:problem b symmetry}.]{
        \includegraphics[width=0.5\columnwidth]{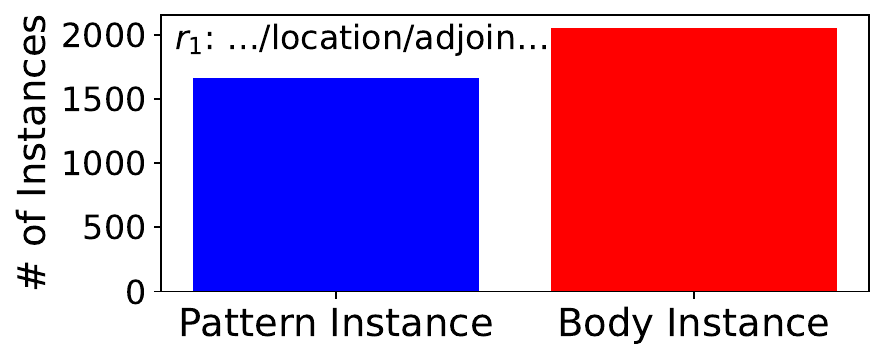}
        \hfill
        \includegraphics[width=0.5\columnwidth]{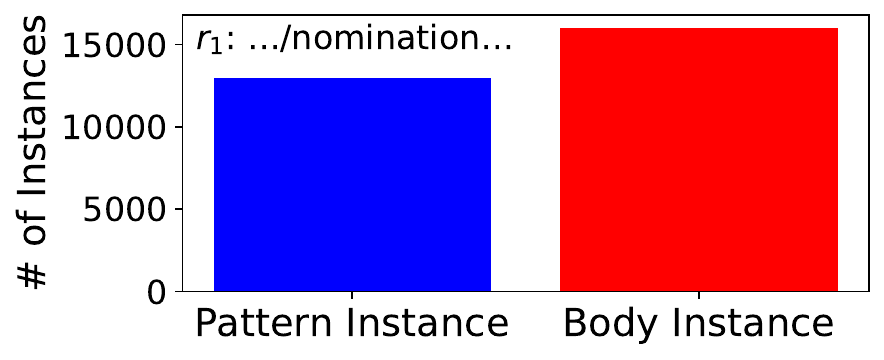}
        \label{fig:PI_BI_universal_symmetry}
    }
    
    \caption{The number of pattern instances and body instances for the local and universal symmetry patterns introduced in Figure~\ref{fig:problem_symmetry}}
    \label{fig:PI_BI_symmetry}
\end{figure}

\paragraph{Empirical Evidence for the Symmetry Pattern} To demonstrate that the over-generalization effect is not limited to the composition pattern discussed in Section~\ref{sec:problem}, we conduct an additional analysis on the symmetry pattern using the same framework in Section~\ref{sec:problem}, as follows.

Similar to the composition pattern, PairRE also induces a relation-level pattern condition for symmetry patterns. If a relation $r_1$ is symmetric, PairRE satisfies the following condition:

\begin{equation}
\begin{aligned}
\textstyle (r_1^H)^2 = (r_1^T)^2 
\end{aligned}
\end{equation}

This condition is determined only by relation embeddings. Therefore, once this condition is learned from observed triples, the model can generalize the symmetry pattern to other entities, even when the pattern is supported by only a small number of instances. This can lead to pattern over-generalization for local symmetry patterns. We empirically verify this by measuring the distribution of $\Delta = (r_1^H)^2 - (r_1^T)^2$. Figure~\ref{fig:problem_symmetry} shows the histograms of the embedding difference ($\Delta$) of local and universal symmetry patterns. Figures~\ref{fig:problem a symmetry} and~\ref{fig:problem b symmetry} show that the elements of $\Delta$ are concentrated near zero for both local and universal patterns, indicating that the model recognizes both as valid symmetry patterns regardless of instance frequency. The numbers of pattern instances and body instances for these symmetry patterns are shown in Figure~\ref{fig:PI_BI_symmetry}. This analysis empirically shows that pattern over-generalization also occurs in another pattern type.

\begin{table*}[t]
\scriptsize
\centering
\begin{tabular}{llccccc}
\hline
\multirow{2}{*}{Model} & \multicolumn{1}{c}{\multirow{2}{*}{Score Function}} & \multicolumn{1}{c}{\multirow{2}{*}{Pattern Modeling}} & \multicolumn{1}{c}{\multirow{2}{*}{Over-Generalization}} & \multicolumn{1}{c}{\multirow{2}{*}{Scalability}} & \multicolumn{2}{c}{MRR}                                    \\
                       & \multicolumn{1}{c}{}                                & \multicolumn{1}{c}{}                                  & \multicolumn{1}{c}{}                                     & \multicolumn{1}{c}{}                             & \multicolumn{1}{c}{WN18RR} & \multicolumn{1}{c}{FB15k-237} \\ \hline
RERSCAL                &    $<h^TW_rt>$                                                 &                    \ding{55}                                   &     -                                                     &      \ding{55}                                            &    .420                        &  .270                               \\
TransR                 &      $-\|P_rh + r - P_rt \|$                                               &                        \ding{55}                               &      -                                                    &  \ding{55}                                                &   .220                         &  .299                             \\
CompoundE              &     $-\|M_r\cdot h-\hat{M}_r \cdot t \|$                                                 &      \ding{51}                                                 &    \ding{51}                                                      &      \ding{51}                                            &                    .491        &    .357                           \\ \hline
PogRE                   &      $-\|L_r  h_r- t_r\|$                                               &       \ding{51}                                                &        \ding{55}                                                  &                                \ding{51}                  &  \textbf{.506}                          &     \textbf{.369}                          \\ \hline
\end{tabular}
\caption{Summary of differences between PogRE and the existing linear transformation models. RESCAL is reproduced in \citet{wang2019evaluating}, and TransR is reproduced in \citet{zhang2020improve}. $W_r$ and $P_r$ are $\mathbb{R}^{d\times d}$ dense linear transformations, where $d$ is the dimension of entities and relations.}
\label{tab:model_comparison_appendix}
\end{table*}

\section{Bounding Constraint Errors Under Practical Training Dynamics}
\label{appendix:Bounding_constraint_error}

In Section~\ref{sec:theoretical_analysis}, we provide a theoretical analysis showing that a dense linear transformation can address over-generalization. However, in practice, KGE embeddings are optimized with negative sampling and a margin-based loss; therefore, the constraint $Ee_i=0$ is approximate. That is, satisfying the exact linear constraint $Ee_i=0$ is challenging due to the approximate nature of margin-based optimization with negative sampling. However, we can mathematically guarantee that the constraint error for unseen entities is bounded by the constraint error of observed entities. Specifically, let us assume the model is sufficiently trained such that the constraint error is minimized within a small margin $\epsilon$ for the observed linearly independent entities $e_1, \dots, e_d$ (i.e., we consider the practical scenario in which $\|Ee_i\| < \epsilon$, rather than the exact condition $\|Ee_i\| = 0$.) Since we empirically verified that the learned entity embeddings form a basis (Section~\ref{sec:experiment_independent}), any unseen entity $e_{new}$ can be represented as a linear combination of the observed entities: $e_{new} = \sum_{i=1}^{d} c_i e_i$. By the linearity of the transformation $E$ and the triangle inequality, the error for the unseen entity is bounded as follows:
\begin{equation}
\begin{aligned}
\|Ee_{new}\| &= \left\|\sum_{i=1}^{d} c_i(Ee_i)\right\| \\
&\leq \sum_{i=1}^{d} |c_i| \cdot \|Ee_i\| \\
&< \left(\sum_{i=1}^{d} |c_i|\right)\epsilon
\end{aligned}
\end{equation}
This inequality demonstrates that minimizing the constraint error of observed entities ($\epsilon \to 0$) directly suppresses the constraint error for unseen entities. Therefore, even under the approximate optimization of margin-based loss, the constraint error of unseen entities remains bounded. Through this analysis, we clarify the practical scope of the theoretical guarantees of our method. In practice, exact constraint satisfaction, corresponding to the ideal zero-error case, i.e., $\|Ee_i\| = 0$, is not guaranteed after training. Rather, when the constraint errors of observed entities are small, corresponding to the approximate case, i.e., $\|Ee_i\| < \epsilon$, the errors of unseen entities can also be bounded, thereby ensuring the practical effectiveness of our proposed method.

\paragraph{Practical Strength of the Assumptions}
Our theoretical analysis involves two assumptions: (1) the approximate satisfaction of pattern constraints for observed entities and (2) the availability of $d+1$ linearly independent observed entities. We discuss the practical strength of these assumptions below.

First, the practical strength of the approximate pattern-constraint assumption depends on how small the residuals become in practice. The approximate satisfaction of pattern constraints is encouraged by the KGE training objective. The training objective reduces the errors of observed pattern instances, thereby encouraging small residuals in the corresponding pattern constraints. However, because training relies on mini-batch gradient-based optimization, it is difficult to know how small the residuals will be after training, making the practical strength difficult to assess a priori. Consequently, our approximate analysis above is conditional on the residuals actually achieved after training.

Second, the requirement of observing $d+1$ linearly independent entities should be understood as a sufficient condition for universal generalization, rather than as a condition that must always hold. As more linearly independent entities supporting a pattern are observed, the pattern constraint applies to a larger subspace; universal generalization is guaranteed when these entities span the relevant space. We regard this requirement as part of an inherent trade-off. If this condition is made less strict, the model may generalize patterns more easily. However, this may also increase the risk of generalizing weakly supported patterns too broadly, which may lead to the over-generalization problem that PogRE is designed to avoid.

\section{Differences between the existing linear transformation model and PogRE}
\label{appendix:difference_between_linear_models}
We present a comparison between existing linear transformation models and PogRE in Table~\ref{tab:model_comparison_appendix}. Existing models that use dense linear transformation such as RESCAL~\cite{nickel2011three} and TransR~\cite{lin2015learning} are not designed for pattern modeling; moreover, when the entity and relation dimensions are $n$, they assign an $\mathbb{R}^{n \times n}$ matrix to each relation, leading to high computational costs as the dimension of entities and relations increases. While CompoundE is capable of pattern modeling and avoids these computational issues by using sparse affine transformation, it suffers from over-generalization. In contrast, PogRE is designed for pattern modeling, addresses the over-generalization problem, and avoids the computational cost issue using a QR-decomposition-inspired method. In addition, PogRE outperforms existing models on WN18RR and FB15k-237, demonstrating its effectiveness.

\begin{table}[t]
\scriptsize
\centering
\begin{tabular}{lcc}
\hline
Model
& WN18RR
& FB15k-237 \\
\hline

R-GCN~\cite{schlichtkrull2018modeling}
& --
& .248 \\

SACN~\cite{shang2019end}
& .470
& .350 \\

CompGCN~\cite{Vashishth2020Composition-based}
& .479
& .355 \\

MRGAT~\cite{dai2022mrgat}
& .481
& .358 \\

CompGCN-MLP~\cite{li2023message}
& .473
& .355 \\
\hline
PogRE
& \textbf{.506}
& \textbf{.369} \\

\hline
\end{tabular}
\caption{MRR comparison between PogRE and transductive GNN-based models on WN18RR and FB15k-237.}
\label{tab:gnn_comparison}
\end{table}

\section{Comparison with Transductive GNNs}
\label{appendix:gnn_comparison}

Transductive GNN models such as R-GCN~\cite{schlichtkrull2018modeling}, SACN~\cite{shang2019end}, and CompGCN~\cite{Vashishth2020Composition-based} improve entity representations through message passing, which incorporates local structural context into entity embeddings. By leveraging structural context, these models may also alleviate pattern over-generalization. However, pattern over-generalization has not been explicitly discussed or analyzed in transductive GNN models. In particular, prior work has not characterized how transductive GNN models capture patterns or how the patterns captured by transductive GNN models are generalized from observed evidence. In contrast, PogRE explicitly models patterns and is designed to improve entity representations while alleviating pattern over-generalization.

We further compare PogRE with these transductive GNN models in terms of MRR. As shown in Table~\ref{tab:gnn_comparison}, PogRE outperforms the baselines across the available benchmarks, demonstrating that PogRE remains effective compared with graph-contextual baselines.

\begin{table*}[t]
\footnotesize
\renewcommand{\arraystretch}{0.98}
\centering
\setlength{\tabcolsep}{3pt}
\begin{tabular}{cccccccc}
\hline
\multirow{2}{*}{Pattern} & \multicolumn{1}{c}{\multirow{2}{*}{\begin{tabular}[c]{@{}c@{}}\# of Pattern \\ Instances (n)\end{tabular}}} & \multicolumn{2}{c}{WN18RR} & \multicolumn{2}{c}{FB15k-237} & \multicolumn{2}{c}{YAGO3-10} \\ \cline{3-8} 
 & \multicolumn{1}{c}{} & \# of Patterns & Prop. (\%) & \# of Patterns & Prop. (\%) & \# of Patterns & Prop. (\%) \\  \hline
\multirow{6}{*}{Symmetry} & n = 1 & - & - & 6 & 13.6 & - & - \\
 & 1 $<$ n $\leq$ 10 & 1 & 20.0 & 4 & 9.1 & 5 & 41.7 \\
 & 10 $<$ n $\leq 10^2$ & 1 & 20.0 & 10 & 22.7 & - & - \\
 & $10^2 <$ n $\leq 10^3$ & 1 & 20.0 & 17 & 38.6 & 4 & 33.3 \\
 & n $> 10^3$ & 2 & 40.0 & 7 & 15.9 & 3 & 25.0 \\ \cline{2-8} 
 & \textbf{Total} & \textbf{5} & 100\% & \textbf{44} & 100\% & \textbf{12} & 100\% \\
\hline
\hline
\multirow{6}{*}{Antisymmetry} 
 & n = 1 & 1 & 9.1 & 1 & 1.8 & - & - \\
 & 1 $<$ n $\leq$ 10 & 2 & 18.2 & 13 & 22.8 & 3 & 20.0 \\
 & 10 $<$ n $\leq 10^2$ & 2 & 18.2 & 26 & 45.6 & 5 & 33.3 \\
 & $10^2 <$ n $\leq 10^3$ & 2 & 18.2 & 13 & 22.8 & 2 & 13.3 \\
 & n $> 10^3$ & 4 & 36.4 & 4 & 7.0 & 5 & 33.3 \\ \cline{2-8} 
 & \textbf{Total} & \textbf{11} & 100\% & \textbf{57} & 100\% & \textbf{15} & 100\% \\
 
 \hline
 \hline
 
\multirow{6}{*}{Inversion} & n = 1 & 3 & 30.0 & 52 & 18.6 & 6 & 35.3 \\
 & 1 $<$ n $\leq$ 10 & 2 & 20.0 & 86 & 30.7 & 4 & 23.5 \\
 & 10 $<$ n $\leq 10^2$ & 5 & 50.0 & 83 & 29.6 & 4 & 23.5 \\
 & $10^2 <$ n $\leq 10^3$ & - & - & 51 & 18.2 & 1 & 5.9 \\
 & n $> 10^3$ & - & - & 8 & 2.9 & 2 & 11.8 \\ \cline{2-8} 
 & \textbf{Total} & \textbf{10} & 100\% & \textbf{280} & 100\% & \textbf{17} & 100\% \\
 \hline
 \hline
 
\multirow{6}{*}{Composition} & n = 1 & 17 & 48.6 & 1,546 & 26.3 & 56 & 17.6 \\
 & 1 $<$ n $\leq$ 10 & 14 & 40.0 & 2,288 & 39.0 & 112 & 35.2 \\
 & 10 $<$ n $\leq 10^2$ & 4 & 11.4 & 1,439 & 24.5 & 91 & 28.6 \\
 & $10^2 <$ n $\leq 10^3$ & - & - & 489 & 8.3 & 53 & 16.7 \\
 & n $> 10^3$ & - & - & 111 & 1.9 & 6 & 1.9 \\ \cline{2-8} 
 & \textbf{Total} & \textbf{35} & 100\% & \textbf{5,873} & 100\% & \textbf{318} & 100\% \\ 
 \hline
 \hline

\multirow{6}{*}{Hierarchy} 
 & n = 1 & 1 & 12.5 & 72 & 21.4 & 11 & 19.6 \\
 & 1 $<$ n $\leq$ 10 & 2 & 25.0 & 94 & 28.0 & 16 & 28.6 \\
 & 10 $<$ n $\leq 10^2$ & 5 & 62.5 & 114 & 33.9 & 14 & 25.0 \\
 & $10^2 <$ n $\leq 10^3$ & - & - & 50 & 14.9 & 12 & 21.4 \\
 & n $> 10^3$ & - & - & 6 & 1.8 & 3 & 5.4 \\ \cline{2-8} 
 & \textbf{Total} & \textbf{8} & 100\% & \textbf{336} & 100\% & \textbf{56} & 100\% \\
 
 \hline
 \hline
 
\multirow{6}{*}{Intersection} 
 & n = 1 & - & - & 438 & 29.6 & 18 & 15.0 \\
 & 1 $<$ n $\leq$ 10 & - & - & 642 & 43.3 & 54 & 45.0 \\
 & 10 $<$ n $\leq 10^2$ & - & - & 378 & 25.5 & 42 & 35.0 \\
 & $10^2 <$ n $\leq 10^3$ & - & - & 24 & 1.6 & 6 & 5.0 \\
 & n $> 10^3$ & - & - & - & - & - & - \\ \cline{2-8} 
 & \textbf{Total} & \textbf{-} & \textbf{-} & \textbf{1,482} & 100\% & \textbf{120} & 100\% \\
 
 \hline
 \hline
 
\multirow{6}{*}{Transitive} 
 & n = 1 & - & - & 2 & 4.1 & - & - \\
 & 1 $<$ n $\leq$ 10 & 1 & 16.7 & 8 & 16.3 & 5 & 33.3 \\
 & 10 $<$ n $\leq 10^2$ & 2 & 33.3 & 16 & 32.7 & 4 & 26.7 \\
 & $10^2 <$ n $\leq 10^3$ & 3 & 50.0 & 11 & 22.4 & 2 & 13.3 \\
 & n $> 10^3$ & - & - & 12 & 24.5 & 4 & 26.7 \\ \cline{2-8} 
 & \textbf{Total} & \textbf{6} & 100\% & \textbf{49} & 100\% & \textbf{15} & 100\% \\
 
 \hline
 \hline
 
\multirow{6}{*}{G.Intersection} 
 & n = 1 & 2 & 33.3 & 21 & 13.0 & 2 & 18.2 \\
 & 1 $<$ n $\leq$ 10 & - & - & 54 & 33.5 & 5 & 45.5 \\
 & 10 $<$ n $\leq 10^2$ & 4 & 66.7 & 70 & 43.5 & 2 & 18.2 \\
 & $10^2 <$ n $\leq 10^3$ & - & - & 14 & 8.7 & 2 & 18.2 \\
 & n $> 10^3$ & - & - & 2 & 1.2 & - & - \\ \cline{2-8} 
 & \textbf{Total} & \textbf{6} & 100\% & \textbf{161} & 100\% & \textbf{11} & 100\% \\
 
 \hline
 \hline
 
\multirow{6}{*}{B. Transitive} 
 & n = 1 & - & - & 6 & 15.8 & - & - \\
 & 1 $<$ n $\leq$ 10 & - & - & 2 & 5.3 & 1 & 12.5 \\
 & 10 $<$ n $\leq 10^2$ & 1 & 33.3 & 11 & 28.9 & 2 & 25.0 \\
 & $10^2 <$ n $\leq 10^3$ & 2 & 66.6 & 9 & 23.7 & 3 & 37.5 \\
 & n $> 10^3$ & - & - & 10 & 26.3 & 2 & 25.0 \\ \cline{2-8} 
 & \textbf{Total} & \textbf{3} & 100\% & \textbf{38} & 100\% & \textbf{8} & 100\% \\
 
 \hline
 \hline
 
\multirow{6}{*}{B. Composition} 
 & n = 1 & 12 & 40.0 & 1,278 & 28.4 & 3 & 7.1 \\
 & 1 $<$ n $\leq$ 10 & 18 & 60.0 & 1,710 & 38.1 & 15 & 35.7 \\
 & 10 $<$ n $\leq 10^2$ & - & - & 1,053 & 23.4 & 12 & 28.6 \\
 & $10^2 <$ n $\leq 10^3$ & - & - & 345 & 7.7 & 12 & 28.6 \\
 & n $> 10^3$ & - & - & 108 & 2.4 & - & - \\ \cline{2-8} 
 & \textbf{Total} & \textbf{30} & 100\% & \textbf{4,494} & 100\% & \textbf{42} & 100\% \\
 \hline

\end{tabular}

\caption{Distribution of inference patterns: Symmetry/Antisymmetry, Inversion, Composition, Hierarchy, Intersection, Transitive, G.Intersection, B. Transitive and B. Composition according to the number of pattern instances ($n$) across three benchmark datasets.}
\label{tab:integrated_patterns}

\end{table*}

\onecolumn
\small
\setlength{\LTcapwidth}{0.95\textwidth}
\begin{longtable}[c]{c p{6.5cm} p{6.5cm}}
\hline
\# of Pattern Instances & Semantically Universal ($n$) & Semantically Local ($n$) \\
\hline

$n \le 10$ & - & \textit{owns} (4) (S) \\
 & & \textit{isAffiliatedTo} (10) (S) \\
 & & \textit{isKnownFor} (4) (S) \\
 & & \textit{created} (2) (S) \\
 & & \textit{hasAcademicAdvisor} (4) (S) \\
 & & \textit{influences}, \textit{created} (3) (I) \\
 & & \textit{influences}, \textit{isInterestedIn} (8) (I) \\
 & & \textit{participatedIn}, \textit{isCitizenOf} (1) (I) \\
 & & \textit{created}, \textit{hasAcademicAdvisor} (1) (I) \\
 & & \textit{isAffiliatedTo}, \textit{isMarriedTo} (1) (I) \\
 & & \textit{isLocatedIn}, \textit{isConnectedTo} (1) (I) \\
 & & \textit{influences}, \textit{isKnownFor} (5) (I) \\
 & & \textit{isLocatedIn}, \textit{dealsWith} (1) (I) \\
 & & \textit{influences}, \textit{hasChild} (1) (I) \\
 & & \textit{isMarriedTo}, \textit{influences} (8) (I) \\
 & & \textit{\_synset\_domain\_topic\_of} (2) (S) \\
 & & \textit{\_synset\_domain\_topic\_of}, \textit{\_has\_part} (9) (I) \\
 & & \textit{\_hypernym},  \\
 & & \textit{\_derivationally\_related\_form} (17) (I) \\
 & & \textit{\_derivationally\_related\_form}, \\
 & & \textit{\_member\_meronym} (23) (I) \\
 & & \textit{\_hypernym}, \textit{\_also\_see} (38) (I) \\
 & & \textit{\_hypernym}, \textit{\_synset\_domain\_topic\_of} (4) (I) \\
 & & \textit{\_also\_see}, \textit{\_verb\_group} (1) (I) \\
 & & \textit{\_hypernym}, \textit{\_verb\_group} (17) (I) \\
 & & \textit{\_has\_part}, \\
 & & \textit{\_member\_of\_domain\_region} (1) (I) \\
 & & \textit{\_instance\_hypernym}, \\
 & & \textit{\_member\_of\_domain\_region} (1) (I) \\
 & & \textit{\_derivationally\_related\_form}, \textit{\_synset\_domain\_topic\_of} (23) (I) \\
 & & \textit{/location/ ...division/country} (4)  (S) \\
 & & \textit{/location/country/capital} (3) (S) \\
 & & \textit{/film/film/prequel} (2) (S) \\
 & & \textit{/people/person/profession} (1) (S) \\
 & & \textit{/film/film/genre} (1) (S) \\
 & & \textit{/film/film\_subject/films} (1) (S) \\
 & & \textit{/base/aareas/schema} \\
 & & \textit{/administrative\_area/capital} (1) (S) \\
 & & \textit{/medicine/symptom/symptom\_of} (1) (S) \\
 & & \textit{/music/instrument/family} (1) (S) \\
 & & \textit{.../sibling\_relationship} \textit{.../influenced\_by} (2) (I) \\
 & & \textit{../romantic\_relationship/celebrity} (1) (I) \\

\hline

$10<n \le 100$ & \textit{\_similar\_to} (74)  (S) & \textit{isLocatedIn}, \textit{participatedIn} (50) (I) \\
 & & \textit{isMarriedTo}, \textit{hasChild} (59) (I) \\
 & & \textit{influences}, \textit{hasAcademicAdvisor} (63) (I) \\
 & & \textit{isLocatedIn}, \textit{owns} (60) (I) \\

\hline

$100 < n \le 1000$ & \textit{dealsWith} (160) (S) & \textit{influences} (180) (S) \\
 & \textit{hasNeighbor} (550) (S) & \textit{hasChild} (414) (S) \\
 & \textit{\_also\_see} (828) (S) & \textit{dealsWith}, \textit{hasNeighbor} (165) (I) \\
 & \textit{.../legislative\_sessions} (668)  (S) & \textit{.../award\_nomination/nominated\_for} (592) (S) \\
 & \textit{.../military\_combatant\_group/combatants} (620) (S) & \textit{.../recording\ .../performance\_role} (402) (S) \\
 & \textit{.../canoodled/participant} (368) (S) & \textit{.../award\_honor/honored\_for} (392) (S) \\
 & \textit{.../marriage/spouse} (342) (S) & \textit{.../education/major\_field\_of\_study} (282) (S) \\
 & \textit{.../friendship/friend} (204) (S) & \textit{.../sports\_team\_roster/position} (106) (S) \\
 & & \textit{.../location/contains}, \textit{.../first\_level\_division\_of} \\
 & & (124) (I) \\
 & & \textit{.../genre/titles}, \textit{...film/country} (157) (I) \\
 & & \textit{.../performance/film}, \textit{...award\_winner} (963) (I) \\
 & & \textit{.../sports\_team\_roster/team}, 
 
 \textit{.../american\_football/ .../position} (470) (I) \\
 & & \textit{...location/contains}, 
 
 \textit{.../mailing\_address/state\_province\_region} \\
 & & (359) (I) \\
 & & \textit{.../regular\_tv\_appearance/actor}, \textit{.../award\_nomination/nominated\_for} (445) (I) \\
 & & \textit{...location/contains}, 
 
 \textit{.../mailing\_address/country} (103) (I) \\
 & & \textit{.../award\_nomination/nominated\_for}, 
 
 \textit{.../produced\_by} (497) (I) \\
 & & \textit{.../dated/participant}, 
 
 \textit{../romantic\_relationship/celebrity} (136) (I) \\
 & & \textit{.../award\_nomination/nominated\_for}, \textit{/film/film/music} (314) (I) \\
 & & \textit{.../music/group\_membership/role}, 
 
 \textit{.../performance\_role} (246) (I) \\
 & & \textit{.../dated/participant}, \textit{.../spouse} (108) (I) \\

\hline

$n > 1000$ & \textit{isMarriedTo} (3674) (S) & \textit{isLocatedIn} (5742) (S) \\
 & \textit{happenedIn}, \textit{participatedIn} (1468) (I) & \textit{isLocatedIn}, \textit{hasCapital} (1743) (I) \\
 & \textit{\_derivationally\_related\_form} (27701) (S) & \\
 & \textit{\_verb\_group} (1060) (S) & \\
 & \textit{.../award\_nomination/award\_nominee} (12950) (S) & \\
 & \textit{.../award\_honor/award\_winner} (6860) (S) & \\
 & \textit{.../track\_contribution/role} (3068) (S) & \\
 & \textit{.../group\_membership/role} (2170) (S) & \\
 & \textit{.../adjoining\_relationship/adjoins} (1660) (S) & \\
 & \textit{.../friendship/participant} (1216) (S) & \\
 & \textit{.../dated/participant} (1134) (S) & \\

\hline
\caption{Human verification results of semantically universal and semantically local patterns, categorized by the number of pattern instances. The number next to each pattern indicates the number of pattern instances, and (S) and (I) indicate symmetry and inversion, respectively.}
\label{tab:relation_symmetry} \\
\end{longtable}
\twocolumn

\end{document}